\documentclass[journal]{IEEEtran}
\IEEEoverridecommandlockouts
\usepackage{cite}
\usepackage{amsmath,amssymb,amsfonts, amsthm}
\usepackage{algorithmic}
\usepackage[ruled,linesnumbered]{algorithm2e}
\usepackage{graphicx}
\usepackage{textcomp}
\usepackage{stfloats}
\usepackage{url}
\usepackage{verbatim}
\usepackage{xcolor}
\usepackage{hyperref}
\usepackage{diagbox}
\usepackage{makecell}
\usepackage{adjustbox}
\usepackage{booktabs}
\usepackage{subfigure}
\usepackage{multirow}
\usepackage{enumitem}
\usepackage{orcidlink}
\newtheorem{theorem}{Theorem}
\usepackage[normalem]{ulem}
\useunder{\uline}{\ul}{}
\newcommand{\RevisionOne}[1]{{#1}}
\newcommand{\RevisionTwo}[1]{{#1}}
\newcommand{\SynC}[1]{\texttt{SynC}}
\begin{document}

\title{\SynC{}: Synergistic Boosting of Structure and Representation for Deep Graph Clustering}
\author{Shifei~Ding\textsuperscript{\orcidlink{0000-0002-1391-2717}},~\IEEEmembership{Senior~Member,~IEEE,}~Benyu~Wu\textsuperscript{\orcidlink{0009-0007-4225-5924}},~Xiao~Xu\textsuperscript{\orcidlink{0000-0003-2888-7451}},~Ling~Ding\textsuperscript{\orcidlink{0000-0002-3208-2528}},~and~Xindong~Wu\textsuperscript{\orcidlink{0000-0003-2396-1704}},~\IEEEmembership{Fellow,~IEEE}

\thanks{This work was supported by the National Natural Science Foundation of China under Grants 62276265, 62576344 (\textit{Corresponding authors:~Benyu~Wu,~Xiao~Xu}).}

\thanks{Shifei Ding, Benyu Wu, Xiao Xu are with the School of Computer Science and Technology/the School of Artificial Intelligence, China University of Mining and Technology, Xuzhou, 221116 China, and with the Mine Digitization Engineering Research Center of Ministry of Education, Xuzhou, 221116 China (email: dingsf@cumt.edu.cn, bywu109@163.com, xu\_xiao@cumt.edu.cn).}

\thanks{Ling Ding is with the College of Intelligence and Computing, Tianjin University, Tianjin, 300350 China (e-mail: dltjdx2022@tju.edu.cn).}

\thanks{Xindong Wu is with the Key Laboratory of Knowledge Engineering with Big Data (the Ministry of Education of China), Hefei University of Technology, Hefei, 230009 China (e-mail: xwu@hfut.edu.cn).}

\thanks{Digital Object Identifier 10.1109/TNNLS.2025.3643594}

\thanks{2162-237X~\copyright~2025 IEEE}
}
\markboth{IEEE TRANSACTIONS ON NEURAL NETWORKS AND LEARNING SYSTEMS}%
{}



\maketitle

\begin{abstract}
\RevisionOne{Employing graph neural networks (GNNs) for graph clustering has shown promising results in deep graph clustering. However, existing methods disregard the reciprocal relationship between representation learning and structure augmentation: the more homogeneous the graph, the more cohesive the node representations; the more cohesive the node representations, the more reliable the structure augmentation becomes. Moreover, the generalization ability of existing GNN-based models on the low homophily graph is relatively poor. To this end, we propose a graph clustering framework named \underline{Syn}ergistic Deep Graph \underline{C}lustering Network (\SynC{}). \SynC{} employs a Transform Input Graph Auto-Encoder (TIGAE) to obtain high-quality embeddings via mitigating the representations collapse issue of GAE for guiding structure augmentation. Then, we re-capture neighborhood representations on the refined graph to obtain clustering-friendly embeddings and conduct self-supervised clustering. Notably, these two stages share weights, resulting in synergistic boosting while significantly reducing the number of model parameters. Additionally, we introduce a structure fine-tuning strategy to improve the model's generalization on the low homophily graph. Extensive experiments on benchmark datasets demonstrate the superiority of \SynC{}. The code is released at GitHub\footnote{\href{https://github.com/Marigoldwu/SynC}{https://github.com/Marigoldwu/SynC}}.}

\end{abstract}


\section{Introduction}
\begin{figure}[htbp]
  \centering
  \label{fig:heatmap}
  \subfigure[Raw]{
    \includegraphics[trim=70pt 30pt 25pt 30pt,clip, width=0.29\columnwidth]{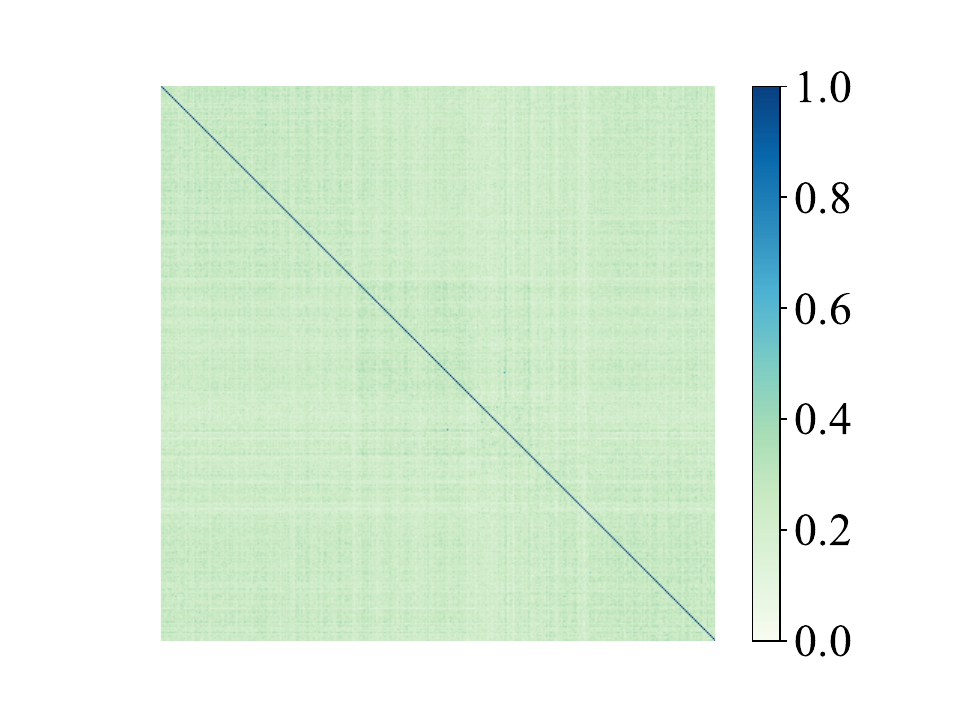}
    \label{fig:raw}
  }
  \hfill
  \subfigure[Layer 1]{
    \includegraphics[trim=70pt 30pt 25pt 30pt,clip,width=0.29\columnwidth]{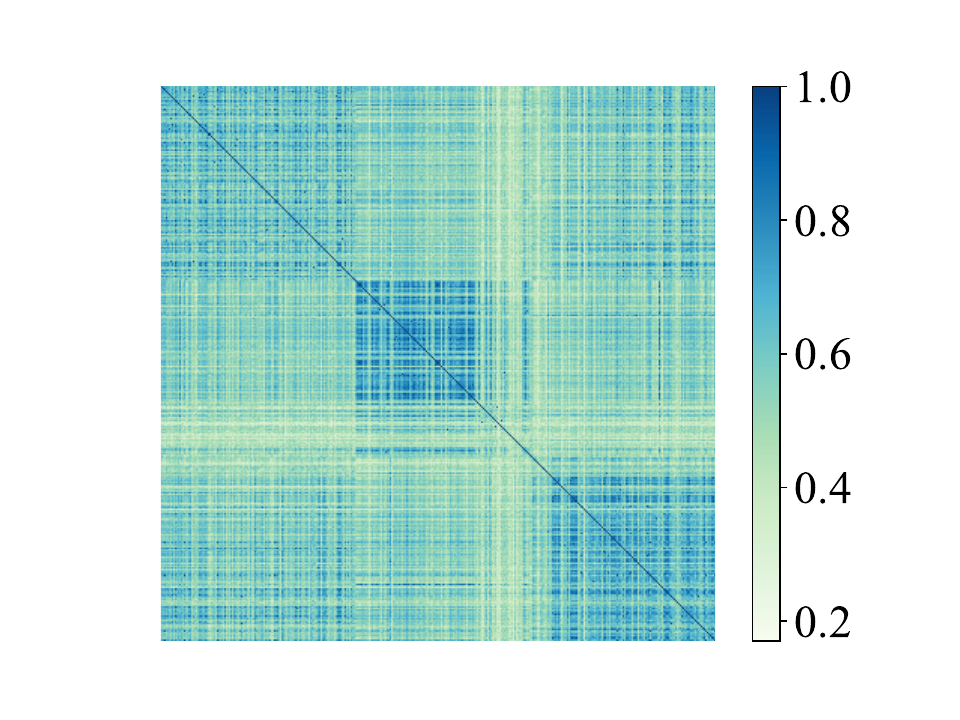}
    \label{fig:gae1}
  }
  \hfill
  \subfigure[Embedding]{
    \includegraphics[trim=70pt 30pt 25pt 30pt,clip,width=0.29\columnwidth]{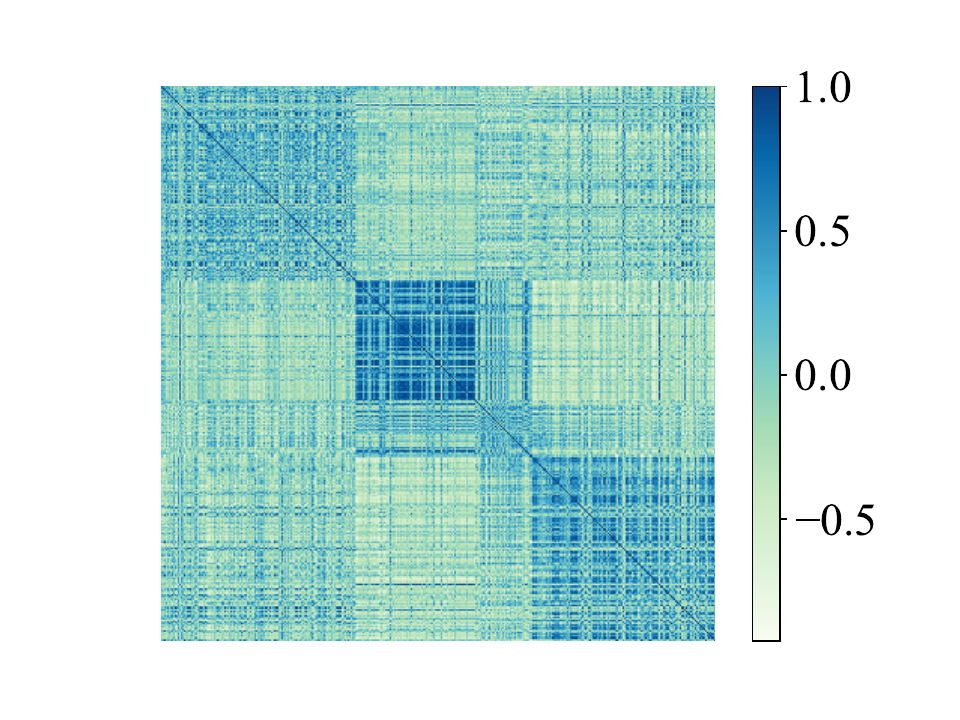}
    \label{fig:gaee}
  }
  
  \subfigure[Enhanced $\mathbf{X}$]{
    \includegraphics[trim=70pt 30pt 25pt 30pt,clip,width=0.29\columnwidth]{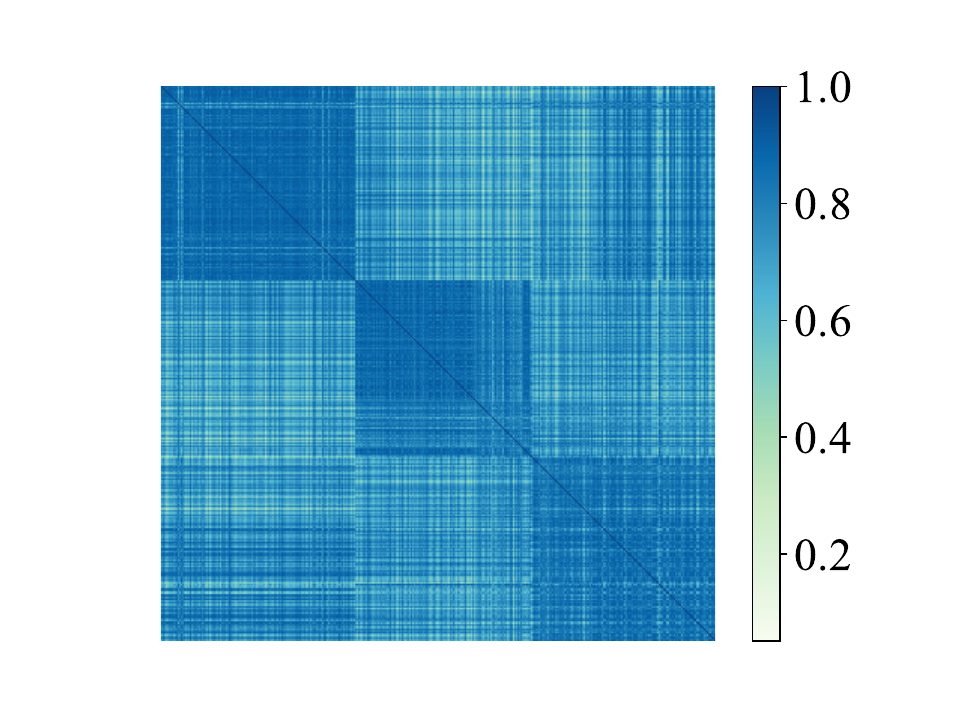}
    \label{fig:xt}
  }
  \hfill
  \subfigure[Layer 1]{
    \includegraphics[trim=70pt 30pt 25pt 30pt,clip,width=0.29\columnwidth]{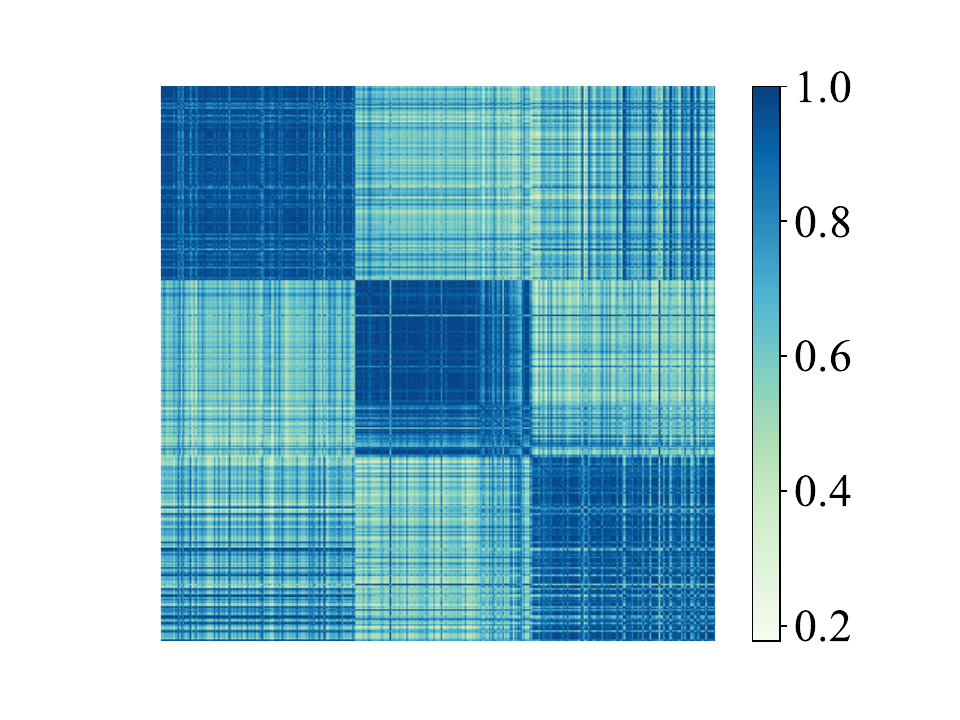}
    \label{fig:TIGAE1}
  }
  \hfill
  \subfigure[Embedding]{
    \includegraphics[trim=70pt 30pt 25pt 30pt,clip,width=0.29\columnwidth]{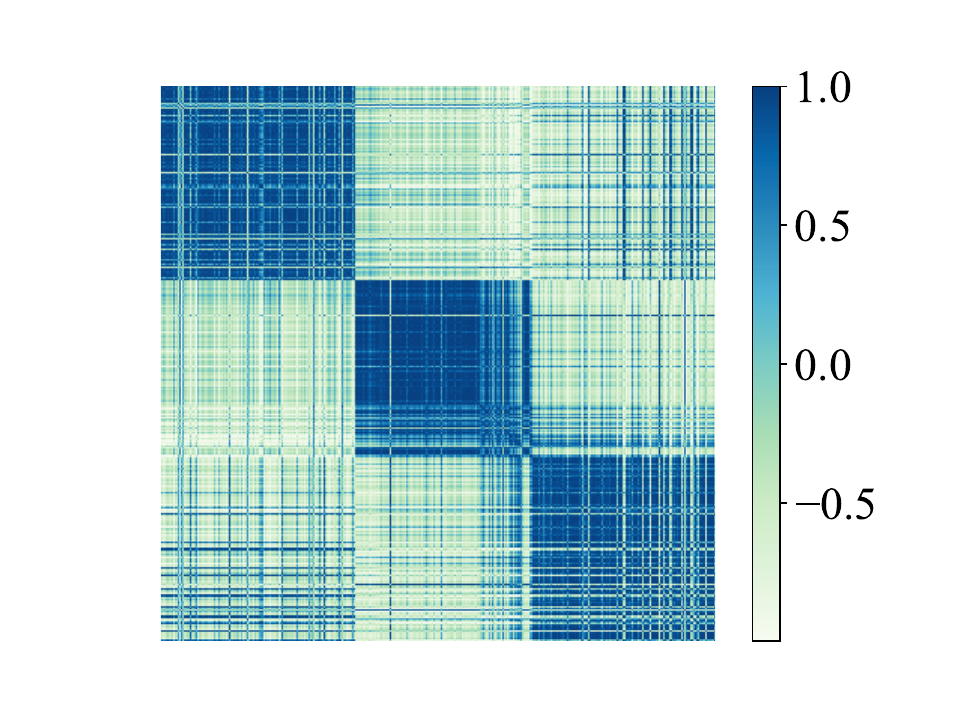}
    \label{fig:TIGAEe}
  }
  \caption{Comparison of node similarity learned by GAE (the first row) and TIGAE (the second row) on the dataset ACM.}
\end{figure}

\IEEEPARstart{W}{ith} the rapid advancement of graph neural networks (GNNs), employing GNNs to learn structural information has garnered significant attention and has yielded notable results in deep graph clustering (DGC)~\cite{hao2023deep, liu2023hard, ding2023graph}. One of the most representative GNNs is the graph convolutional network (GCN)~\cite{kipf2016semisupervised}, including its self-supervised version, the graph auto-encoder (GAE)~\cite{kipf2016variational}, which learns low-dimensional representations of the input. GAE encounters the representation collapse issue that nodes from different categories are mapped to similar representations~\cite{liu2022deep}. Figs.~\ref{fig:raw}-\ref{fig:gaee} illustrate the cosine similarity matrices among nodes on the ACM. It is evident that even after the transformation by GAE, the nodes from the three categories still cannot be well distinguished. Nevertheless, as illustrated in Figs.~\ref{fig:xt}-\ref{fig:TIGAEe}, the final embeddings exhibit a significant improvement in discriminability after performing a linear transformation with cosine similarity preserved. Based on this exciting observation, we infer that the quality of the input attributes plays a crucial role in determining the discriminability of the GAE embeddings.

Moreover, the over-smoothing effect~\cite{li2018deeper} could be regarded as a potential benefit for clustering, particularly when it facilitates intra-cluster smoothing. Zhao et al.~\cite{zhao2021data} have also highlighted that the representations of any nodes within the same class are identical after applying multi-layer GCNs on an ideal graph or a class-homophilic graph. Therefore, GCN-based models typically do not perform well on graphs with low homophily, which is a manifestation of poor generalization. Graph Augmentation Learning (GAL) can improve the generalization of graph models and has been widely studied. Using neural networks to predict graph structures~\cite{zhao2021data} is an interesting attempt. However, this method often uses inconsistent networks between graph prediction and downstream tasks, complicating model optimization. \RevisionOne{Graph structure and downstream tasks are closely related. For example, in graph clustering, high-homophily graphs facilitate learning cohesive node representations. Using separate networks here overlooks the reciprocal relationship between representation learning and structure augmentation. Fortunately, in unsupervised tasks, both relate to node embedding: high-quality embedding helps reconstruct the original graph and improve its homogeneity, which in turn facilitates learning better representations.}

\RevisionOne{To this end, we propose a novel graph clustering framework termed \underline{Syn}ergistic Deep Graph \underline{C}lustering Network (\SynC{}). \SynC{} introduces the Transform Input GAE (TIGAE) to optimize embeddings. TIGAE-generated high-quality embeddings first predict inter-node edge probabilities; a structure fine-tuning strategy (integrating pruning, linking, and weight assignment) then refines the predicted graph precisely. This refined graph aids feature aggregation, laying the groundwork for better generalization. Critically, the refined graph is fed into another weight-sharing TIGAE to obtain improved embeddings, forming a deeply synergistic mechanism: the first TIGAE’s embeddings provide an initial basis for graph prediction and refinement, and when the refined graph is fed back, weight sharing synchronously updates the first TIGAE’s parameters, creating synergy where embeddings guide graph optimization and the optimized graph enhances embeddings. Finally, clustering is achieved via self-supervised clustering on these embeddings. The main contributions of this paper are summarized as follows.}

\noindent \textbf{i}) We propose \SynC{}, which leverages high-quality embeddings from TIGAE to achieve synergistic boosting between feature learning and structure optimization, enhancing intra-cluster similarity and inter-cluster discriminability.\\
\textbf{ii}) We develop an improved GAE variant for DGC, namely TIGAE. It optimizes the input of GCN by indirectly integrating explicit structural information, thus mitigating the representation collapse issue of vanilla GAE.\\
\textbf{iii}) We design a multi-factor structure fine-tuning strategy that refines the predicted graph precisely, laying a solid foundation for enhancing the model’s generalization performance.

The rest of the paper is organized as follows: Section~\ref{sec:related_work} introduces related work on DGC and GAL, respectively. Section~\ref{sec:methodology} details our proposed \SynC{}. Section~\ref{sec:experiments} analyzes experimental results. Section~\ref{sec:conclusion} concludes with an outlook.

\section{Related Work}
\label{sec:related_work}
\subsection{Deep Graph Clustering}
DGC aims to cluster nodes into distinct groups via node embeddings, ensuring higher intra-cluster than inter-cluster similarity. Earlier methods learned representations solely by mapping attributes to low-dimensional embeddings and then conducted clustering using K-Means~\cite{hartigan1979algorithm} or self-supervised methods (e.g., DEC~\cite{xie2016unsupervised}). 

Graph representation learning (GRL) has advanced notably since GCN, fueling DGC research. Techniques like GAE and GAT~\cite{velivckovic2018graph} enhance performance via more cohesive embeddings, with various GAE variants (e.g., MGAE~\cite{wang2017mgae}, MaskGAE~\cite{li2023what}, GraphMAE2~\cite{hou2023graphmae2}). SDCN~\cite{bo2020structural} highlights joint learning of structure and attributes to enrich embeddings, but lacks dynamic fusion and reliable self-supervised targets. DFCN~\cite{tu2021deep} improves via dynamic cross-modal fusion and triple self-supervised learning to address these. CaEGCN~\cite{huo2021caegcn} uses cross-attention to fuse auto-encoder and GAE representations. AGC and IAGC~\cite{zhang2023adaptive} employ adaptive GCN to leverage high-order structure for DGC.

While cohesion is key for clustering, discriminability is also critical to performance~\cite{ding2024towards}. DCRN~\cite{liu2022deep} boosts embedding discriminability via dual-level information correlation reduction. AGC-DRR~\cite{gong2022attributed} reduces redundancy in input and latent spaces for effective sample discrimination. R$^2$FGC~\cite{yi2023redundancy} captures intrinsic relational and semantic information of non-iid nodes to cut redundancy and get discriminative embeddings. Despite their strong performance, these methods are increasingly complex. Notably, contrastive learning-based graph clustering has advanced with refined positive and negative (including hard) sample selection~\cite{zhao2021graph, yang2023ccgc, liu2023hard}, as have multi-view clustering methods~\cite{zhan2018graph, liu2022stationary, liao2022view}. \RevisionOne{SCGC~\cite{liu2024simple} proposes a cross-view structural consistency objective to optimize a simple multi-layer perceptron network. IDCRN~\cite{liu2025improved} enhances the identity matrix by approximating the cross-view sample correlation matrix to a clustering-refined affinity matrix.} \RevisionTwo{NS4GC~\cite{liu2024reliable} estimates the ideal node similarity matrix through node-neighbor alignment and semantic-aware sparsification, thereby addressing the insufficient exploration of node similarity in graph clustering. HomoCAGC~\cite{chen2025homophily} uses graph homophily to adjust the positive node pairs of the contrastive regularizer through feature-oriented node pairwise similarity. NeuCGC~\cite{peng2025trustworthy} adapts graph homogeneity with dynamic neutral pairs, optimizes neighbor contrast to enhance clustering robustness.}

Progress has also been made in large-scale graph clustering~\cite{liu2023dink} and clustering with unknown cluster numbers~\cite{liu2023reinforcement}. \RevisionOne{CDC~\cite{kang2024cdc} is a simple linear-complexity framework for clustering diverse data.} Notably, heterogeneous graph clustering remains challenging: Shen et al.~\cite{shen2025heterophily} proposed a latent graph-guided unsupervised method to mitigate semantic heterophily. For noisy multiplex graphs, Shen et al.~\cite{shen2024beyond} introduced an information-aware graph structure refinement approach. In subgraph learning, Yang et al.~\cite{yang2023extract} proposed P2GNN to avoid local topology distortion from node sampling, using subgraph-centric pivots that balance independence and regional correlations. For non-graph data, Li et al.~\cite{li2021adaptive} proposed an adaptive graph generation framework, whereas Zhang et al.~\cite{zhang2022non} utilized generative models and anchors to construct bipartite graphs, reducing the complexity of graph convolution. Despite significant advances, DGC still faces challenges in graph data quality, embedding discriminability, model stability, and scalability~\cite{wu2025dgcbench}. For more methods, refer to Liu et al.’s review \cite{liu2023survey} and their \href{https://github.com/yueliu1999/Awesome-Deep-Graph-Clustering}{Github Repository}.

\subsection{Graph Augmentation Learning}
GAL aims to improve GRL and model robustness, notably for incomplete or noisy data~\cite{yu2022graph}. Representative micro-level augmentation methods include DropGNN \cite{papp2021dropgnn} (random node removal) and DropEdge~\cite{rong2020dropedge} (random edge removal), though excessive indiscriminate pruning may limit performance gains.

Additionally, leveraging latent graph links enhances performance: GAUG~\cite{zhao2021data} uses neural networks for edge prediction, with the predicted graph applied to downstream tasks; EGRC-Net~\cite{peng2022graph} adds the most similar nodes to each node via cosine similarity of embeddings; PASTEL~\cite{sun2022position} combines position encoding with node features, using class-wise conflict metrics as edge weights for learning; DFCN-RSP~\cite{gong2022deep} employs random walks to filter noisy edges and strengthen reliable ones via local structural similarity. \RevisionOne{Xu et al.~\cite{xu2025does} theoretically proved edge deletion’s superiority over addition via the Error Passing Rate.} Though effective, these methods either rely on simplistic augmentation guidance or use complex modules.

We propose using the representation learning network for structure augmentation, enabling synergy between representation and structure. Combined with a structure fine-tuning strategy to obtain refined graphs, our approach leverages their mutual enhancement to learn clustering-friendly embeddings.

\begin{figure}[!t]
\centering
\includegraphics[width=\columnwidth]{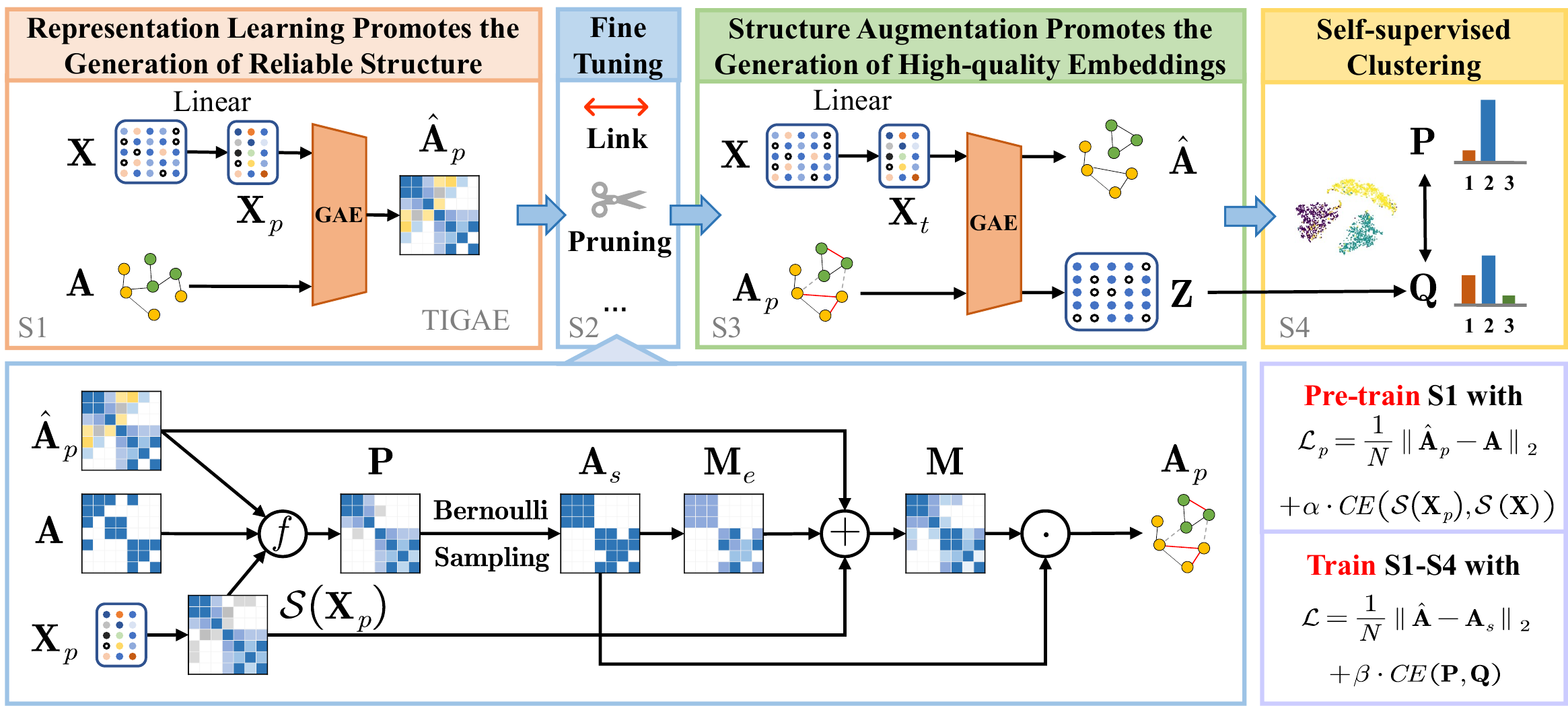}
\caption{\RevisionOne{Overview of our \SynC{}: TIGAE integrates linear transformation and GAE. First, TIGAE is pretrained for initial weights (S1), then generates a gradient-free predicted graph (inference S1). Next, structure fine-tuning prunes, connects, and weights edges in this graph (S2). The refined graph then feeds into TIGAE (weight-sharing with S1) to learn neighbor information, yielding gradient-based final embeddings (S3), enabling synergistic representation learning and structure augmentation. Finally, self-supervised clustering uses these cohesive, discriminative embeddings (S4).}}
\label{fig:model}
\end{figure}

\section{Methodology}
Figure \ref{fig:model} illustrates the schematic framework of \texttt{SynC}.
\label{sec:methodology}

\subsection{Notation Definition}
Consider an undirected graph $\mathcal{G}$ with $\mathcal{E}$ edges, $N$ nodes, and $k$ node categories. Its connectivity is described by the adjacency matrix $\mathbf{A} \in \mathbb{R}^{N \times N}$. Let $\widetilde{\mathbf{A}} = \mathbf{A} + \mathbf{I}$, where $\mathbf{I}$ is the identity matrix. $\mathbf{L}=\mathbf{D}-\mathbf{A}$ is the graph Laplacian matrix. The symmetrically normalized adjacency matrix is $\widetilde{\mathbf{L}} = \widetilde{\mathbf{D}}^{-1/2}\widetilde{\mathbf{A}}\widetilde{\mathbf{D}}^{-1/2}$, with $\widetilde{\mathbf{D}}$ as the degree matrix. Node attributes are given by $\mathbf{X} \in \mathbb{R}^{N \times d}$ ($d$ is the attribute dimension), with linear transforms denoted $\mathbf{X}_t$ and $\mathbf{X}_p$ (subscripts $t, p$ distinguish transform stages). Cosine similarity is given by $\mathcal{S}(\mathbf{X}) = \frac{\mathbf{X}}{\|\mathbf{X}\|_2} \cdot \frac{\mathbf{X}^\top}{\|\mathbf{X}^\top\|_2}$, where $\|\cdot\|_2$ is $L_2$-norm. Superscript $l$ indicates the $l$-th iteration. $\mathcal{E}(\mathbf{F})=1/2 \rm{tr}(\mathbf{F}^\top\mathbf{LF})$ denotes Dirichlet energy of graph signal $\mathbf{F}$. $\hat{\mathbf{A}}_p\in \mathbb{R}^{N \times N}$ is edge probability predicted in the second phase.



\subsection{Transform Input Graph Auto-Encoder}
\label{sec:TIGAE}
GAE employs two GCN layers for encoding attributes and structures. It decodes the embeddings by computing inner products, which can be described as follows.
\begin{equation}
    \mathbf{H}=\widetilde{\mathbf{L}}\text{ReLU}(\widetilde{\mathbf{L}}\mathbf{X}\mathbf{W}_1)\mathbf{W}_2, \hat{\mathbf{A}}=\text{Sigmoid}(\mathbf{HH}^\top).
    \label{eq:gae}
\end{equation}
Here, $\hat{\mathbf{A}}$ is the reconstructed adjacency matrix, and ReLU is the activation function.
To build TIGAE, we add a linear transformation layer with bias before GAE to mitigate representation collapse. This simple transformation offers two benefits: it enables dimension reduction to adjust the feature space for effective information propagation, and its bias matrix indirectly incorporates explicit structure information, enhancing cohesion and discriminability. TIGAE’s formal expression is:
\begin{equation}
    \begin{split}
        \mathbf{Z}&=\widetilde{\mathbf{L}}\text{ReLU}(\widetilde{\mathbf{L}}(\mathbf{X}\mathbf{W}_a^\top+\mathbf{W}_b)\mathbf{W}_1)\mathbf{W}_2, \\
        \hat{\mathbf{A}}_p&=\rm{Sigmoid}(\mathbf{ZZ}^\top).
    \end{split}
    \label{eq:TIGAE}
\end{equation}
In what follows, $\mathbf{X}_p$ or $\mathbf{X}_t$ denote $\mathbf{X}\mathbf{W}_a^\top+\mathbf{W}_b$. Without ReLU activation, $\mathbf{Z}$ in Eq. (\ref{eq:TIGAE}) is as follows:
\begin{equation}
    \begin{split}
\mathbf{Z}&=\widetilde{\mathbf{L}}^2\mathbf{X}\mathbf{W}_a^\top\mathbf{W}_1\mathbf{W}_2+\widetilde{\mathbf{L}}^2\mathbf{W}_b\mathbf{W}_1\mathbf{W}_2.
    \end{split}
    \label{eq:TIGAE_wo}
\end{equation}
The second term in Eq. (\ref{eq:TIGAE_wo}) equals $\widetilde{\mathbf{L}}(\widetilde{\mathbf{L}}\mathbf{W})\mathbf{W}_2$ (where $\mathbf{W}=\mathbf{W}_b\mathbf{W}_1$), i.e., linearly transformed $\widetilde{\mathbf{L}}$ propagated on the original graph. Thus, compared to GAE, TIGAE’s encoder integrates richer information into final embeddings, yielding more accurate graph reconstruction during decoding. \RevisionOne{Moreover, we present the following theorem from the Dirichlet energy perspective \cite{cai2020note}.}

\begin{theorem}[\RevisionOne{Smoothness of Graph Convolution on Low-Dimensional Features}]
\RevisionOne{Given a graph $\mathcal{G}$ and its Laplacian matrix $\mathbf{L}=\mathbf{D}-\mathbf{A}$, for any high-dimensional input features $\mathbf{X}\in\mathbb{R}^{N\times d}$ and a low-dimensional projection matrix $\mathbf{W}_a\in\mathbb{R}^{k\times d}(k<d)$, the low-dimensional embedding $\mathbf{Z}=\widetilde{\mathbf{L}}^2\mathbf{X}\mathbf{W}_a^\top\mathbf{W}$ generated by the GCN has lower Dirichlet energy than the unprojected embedding $\widetilde{\mathbf{L}}^2\mathbf{X}\mathbf{W}$, i.e., $\mathcal{E}(\mathbf{Z})\le\mathcal{E}(\mathbf{H})$, where $\widetilde{\mathbf{L}}$ is the symmetrically normalized graph Laplacian matrix, and $\mathbf{W}=\mathbf{W}_1\mathbf{W}_2$ is the network weight matrix.}
\label{theorem:1}
\end{theorem}

\begin{proof}
\RevisionOne{The Dirichlet energy of the GAE embedding is:}
\RevisionOne{\begin{equation}
    \mathcal{E}_{G}=\frac{1}{2}\rm{tr}(\mathbf{H}^\top\mathbf{L}\mathbf{H})=\frac{1}{2}\rm{tr}((\widetilde{\mathbf{L}}^2\mathbf{X}\mathbf{W})^\top\mathbf{L}(\widetilde{\mathbf{L}}^2\mathbf{X}\mathbf{W})),
\end{equation}}
\RevisionOne{Since the bias term of the linear transformation does not contribute to the energy ($\mathbf{L} \cdot \mathbf{1}_n = \mathbf{0}$), the Dirichlet energy of the TIGAE embedding is:
\begin{equation}
        \mathcal{E}_{T}=\frac{1}{2}\rm{tr}(\mathbf{Z}^\top\mathbf{L}\mathbf{Z})=\frac{1}{2}\rm{tr}((\widetilde{\mathbf{L}}^2\mathbf{X}\mathbf{W}_a^\top\mathbf{W})^\top\mathbf{L}(\widetilde{\mathbf{L}}^2\mathbf{X}\mathbf{W}_a^\top\mathbf{W})),
\end{equation}
\RevisionOne{The energy difference between TIGAE and GAE is:}
\begin{equation}
    \begin{split}
        \Delta\mathcal{E}&=\mathcal{E}_G-\mathcal{E}_T\\
        &=\frac{1}{2}\rm{tr}(\mathbf{L}\widetilde{\mathbf{L}}^2\mathbf{XWW}^\top\mathbf{X}^\top{\widetilde{\mathbf{L}}}^{2^\top})\\&-\frac{1}{2}\rm{tr}(\mathbf{L}\widetilde{\mathbf{L}}^2\mathbf{X}\mathbf{W}_a^\top\mathbf{WW}^\top\mathbf{W}_a\mathbf{X}^\top{\widetilde{\mathbf{L}}}^{2^\top})\\
        &=\frac{1}{2}\rm{tr}(\mathbf{L}\widetilde{\mathbf{L}}^2\mathbf{XW}(\mathbf{I}-\mathbf{W}_a^\top\mathbf{W}_a)\mathbf{W}^\top\mathbf{X}^\top{\widetilde{\mathbf{L}}}^{2^\top}),
    \end{split}
\end{equation}}
\RevisionOne{By the cross-entropy constraint in Eq. (\ref{eq:loss_TIGAE}), $\mathbf{X}_p$ is forced to preserve the direction of $\mathbf{X}$, thus $\mathbf{I} - \mathbf{W}_a^\top \mathbf{W}_a$ is positive semi-definite. Since both the graph Laplacian matrix $\mathbf{L} = \mathbf{D} - \mathbf{A}$ and $\widetilde{\mathbf{L}}^2$ are positive semi-definite matrices, we have $\Delta\mathcal{E}\ge0$.
The smaller the Dirichlet energy $\mathcal{E}$, the smaller the variation of the graph signal between adjacent nodes, i.e., the smoother the signal. Therefore, the low-dimensional projection makes the embedding $\mathbf{Z}$ smoother than the original embedding $\mathbf{H}$.
\textbf{Q.E.D.}}
\end{proof}

To prevent arbitrary transformations, we introduce a cross-entropy loss $CE(\cdot,\cdot)$ between the node similarity of $\mathbf{X}$ and $\mathbf{X}_p$. TIGAE is optimized via the following loss function:
\begin{equation}
    \mathcal{L}_p=\frac{1}{N}\|\hat{\mathbf{A}}_p-\widetilde{\mathbf{A}}\|_2+\alpha\cdot CE(\mathcal{S}(\mathbf{X}_p),\mathcal{S}(\mathbf{X})),
    \label{eq:loss_TIGAE}
\end{equation}
where $\alpha$ is a trade-off hyper-parameter.
\subsection{Structure Fine-tuning Strategy}
While predicting the graph via sampling from the reconstructed graph is feasible, relying solely on edge prediction probabilities and the original graph to generate the predicted graph may introduce biases. This is because the probabilities derive from embeddings generated by GCN on the unreliable original graph. Thus, we propose a structure fine-tuning strategy to refine the graph by integrating multiple factors, facilitating better GRL. We first compute the cosine similarity of $\mathbf{X}_p$ and average it with edge prediction probabilities:
\begin{equation}
\mathbf{P}'=(\hat{\mathbf{A}}_p^{(l)} + \mathcal{S}(\mathbf{X}_p^{(l)})) / 2.
\label{eq:sx}
\end{equation}
$\mathcal{S}(\mathbf{X}_p^{(l)})$ shows more distinct differences than $\mathcal{S}(\mathbf{X}^{(l)})$. \RevisionOne{Cosine similarity can be replaced by Euclidean/Manhattan distance-based similarity or Pearson correlation coefficient, selected based on data type.} We add high-probability edges that are missing from the original graph:
\begin{equation}
\begin{split}
    \mathbf{P}_r'&=\mathbf{P}'-\mathbf{P}'\odot\widetilde{\mathbf{A}}, \\
    \mathbf{A}_{\text{add}_{ij}}^{(l)}=\mathbf{A}_{\text{add}_{ji}}^{(l)}&=1\;\textbf{if}\;\mathbf{P}_{r_{ij}}'==\max{(\mathbf{P}}_i')\;\textbf{else}\;0, \\
    \mathbf{A}_{\text{mask}}^{(l)}&=(\widetilde{\mathbf{A}}+\mathbf{A}_{\text{add}}^{(l)}).
\end{split}
\label{eq:mask}
\end{equation}
Here, $\odot$ denotes the Hadamard product. Next, Bernoulli sampling is introduced to generate an adjacency matrix according to the modified edges probabilities:
\begin{equation}
\begin{split}
    \mathbf{P}^{(l)}&=\mathbf{P}'\odot\mathbf{A}_{\text{mask}}^{(l)}, \\
    \mathbf{A}_{s_{ij}}^{(l)}&=\textbf{Bernoulli}(p=\mathbf{P}_{ij}^{(l)}).
\end{split}
\label{eq:bernoulli}
\end{equation}
Notably, the sampled matrix may be asymmetrical, violating undirected graph prerequisites. To address this, we assign a default weight of $0.5$ to unidirectional edges as a penalty:
\begin{equation}
    \mathbf{A}_s^{(l)}=(\mathbf{A}_s^{(l)}+\mathbf{A}_s^{(l)\top})/2.
\label{eq:as}
\end{equation}
Edges connecting high-degree nodes are more critical. We compute the edge importance matrix $\mathbf{M}_e$ via the degree matrix:
\begin{equation}
    \mathbf{M}_e^{(l)}=L_2({\widetilde{\mathbf{D}}_s}{\mathbf{A}_s^{(l)}}{\widetilde{\mathbf{D}}_s}),
\label{eq:imp}
\end{equation}
where $\widetilde{\mathbf{D}}_s$ is the degree matrix of $\mathbf{A}_s^{(l)}$, and $L_2$ denotes $L_2$ normalization. Finally, we integrate edge prediction probabilities, cosine similarities, and edge importance to obtain a more accurate edge confidence matrix $\mathbf{M}$. The graph refined via our structure fine-tuning strategy is calculated as:
\begin{equation}
\begin{split}
    \mathbf{M}^{(l)}&=(\hat{\mathbf{A}}_p^{(l)} + \mathcal{S}(\mathbf{X}_p^{(l)}) + \mathbf{M}_e^{(l)})/3, \\
    \mathbf{A}_p^{(l)}&=\mathbf{M}^{(l)}\odot\mathbf{A}_s^{(l)}.
\end{split}
\label{eq:a_aug}
\end{equation}
\subsection{Synergistic Interaction Architecture}
GRL on neural network-generated predicted graphs is promising. However, using distinct GNNs for graph prediction and downstream tasks complicates the model. Additionally, edge probability sampling in graph generation involves non-differentiable operations that hinder backpropagation.

To resolve these issues, we propose a novel synergistic architecture of representation learning and structure augmentation, leveraging TIGAE’s strengths in high-quality embedding learning and graph reconstruction. Specifically, we first pretrain TIGAE for reliable weights (ensuring a robust initial predicted graph), then perform gradient-free forward propagation with these weights to obtain edge prediction probabilities. The process is described as:
\begin{equation}
\begin{split}
    \mathbf{Z}_p^{(l)}&=\widetilde{\mathbf{L}}^{(l)}\text{ReLU}(\widetilde{\mathbf{L}}^{(l)}\mathbf{X}_p^{(l)}\mathbf{W}_1^{(l-1)})\mathbf{W}_2^{(l-1)}, \\
    \hat{\mathbf{A}}_p^{(l)}&={\rm{Sigmoid}}(\mathbf{Z}_p^{(l)}{\mathbf{Z}_p^{(l)}}^\top),
\end{split}
\label{eq:prediction}
\end{equation}
where $\mathbf{W}_1^{(0)}$ and $\mathbf{W}_2^{(0)}$ denote pretrained weights. Next, we use structure fine-tuning to refine the predicted graph. The augmented graph then serves as input to TIGAE, where we re-run forward propagation with gradient computation, enabling iterative TIGAE optimization via gradient descent. The process is expressed as:
\begin{equation}
\begin{split}
    \mathbf{Z}^{(l)}&=\widetilde{\mathbf{L}}_p^{(l)}\text{ReLU}(\widetilde{\mathbf{L}}_p^{(l)}\mathbf{X}_t^{(l)}\mathbf{W}_1^{(l)})\mathbf{W}_2^{(l)}, \\
    \hat{\mathbf{A}}^{(l)}&={\rm{Sigmoid}}(\mathbf{Z}^{(l)}{\mathbf{Z}^{(l)}}^\top). \\
\end{split}
\label{eq:clustering}
\end{equation}
$\mathbf{\widetilde{L}}_p^{(l)}$ denotes the symmetric normalized adjacency matrix of the augmented graph $\mathbf{A}_p^{(l)}$. Backward propagation is then performed based on the loss function (detailed in the next section) to optimize the weights.

A single TIGAE enabling synergistic representation learning and structure augmentation greatly reduces parameters and resolves sampling-induced backward propagation issues. Shared weights allow TIGAE to effectively augment structure and learn cohesive representations. \RevisionOne{On one hand, weight sharing facilitates synergistic learning of consistent representations; on the other, optimizing a single component reduces model size and eliminates the need for Gumbel-Softmax reparameterization.} Theoretically, our synergistic framework cuts parameters by 50\% compared with two TIGAEs with separate weights.

\subsection{Self-Supervised Clustering}
Synergistic representation learning and structure augmentation yield final embeddings with strong cohesiveness and discriminability. While direct K-Means on embeddings performs well, we use a self-supervised approach~\cite{van2008visualizing, xie2016unsupervised, bo2020structural} for improved clustering performance. The soft clustering assignment distribution $\mathbf{Q}$ is computed as follows:
\begin{equation}
    {q}_{ij}=\frac{(1+\| \mathbf{z}_i-\boldsymbol{\mu}_j \|^2)^{-1}}{\sum_{j'}(1+\| \mathbf{z}_i - \boldsymbol{\mu}_{j'}\|^2)^{-1}},
    \label{eq:q}
\end{equation}
where $\boldsymbol{\mu}_j$ denotes the $j$-th cluster’s center vector, initialized via K-Means on pre-trained TIGAE embeddings. The self-supervised target distribution $\mathbf{P}$ is computed as:
\begin{equation}
    {p}_{ij}=\frac{{q^2}_{ij}/\sum_i{q}_{ij}}{\sum_{j'}{q^2}_{ij'}/\sum_i{q}_{ij'}}.
    \label{eq:p}
\end{equation}
The loss function combining reconstruction and clustering loss optimizes the framework iteratively:
\begin{equation}
    \mathcal{L}=\frac{1}{N}\|\hat{\mathbf{A}}-\mathbf{A}_s\|_2+\beta\cdot CE(\mathbf{P},\mathbf{Q}).
    \label{eq:loss_total}
\end{equation}
$\beta$ is a hyperparameter. For each node, cluster assignment is determined by the maximum $q_{\cdot j}$. The clustering label $\mathbf{Y}$ is: 
\begin{equation}
    y_i = \arg\max_{j} q_{ij}, \quad j=1,2,...,k.
    \label{eq:y}
\end{equation}
Algorithm \ref{alg:train} shows the complete training process of \SynC{}. 

\subsection{Computational and Space Complexity Analysis}
For the proposed TIGAE, its linear transformation-induced computational complexity is $\mathcal{O}(Ndc)$, where $c$ is the linear transformation dimension. For $d=2c$, GAE (with two GCN computations) has an additional complexity of $\mathcal{O}(N(N+d_1)c)$ compared to TIGAE, where $d_1$ is the output embedding dimension after the first GCN. Thus, TIGAE is more efficient when $N+d_1>d$. Regarding the fine-tuning strategy, its main complexity stems from Eq. (\ref{eq:imp}), which originally reaches $\mathcal{O}(N^3)$ and limits scalability. \RevisionOne{However, $\widetilde{\mathbf{D}}_s$ and $\mathbf{A}_s^{(l)}$ in Eq. (\ref{eq:imp}) are sparse matrices, so matrix compression reduces their complexity to $\mathcal{O}(\mathcal{E}N)$.} The space complexity of \SynC{} is dominated by dense adjacency and similarity matrices, leading to $\mathcal{O}(N^2)$. This is a key barrier to large-scale graph scalability. \RevisionOne{While sampling is widely used for large-scale GNN training \cite{Zeng2020GraphSAINT, chiang2019cluster}, it is incompatible with our model: TIGAE relies on global structure reconstruction (sampling only supports local reconstruction, degrading performance), and structural fine-tuning involves global pruning and linking.}

\begin{algorithm}[t]
\caption{The clustering process of \SynC{}.}
\label{alg:train}
\SetKwInOut{KwIn}{Input}
\KwIn{Attribute $\mathbf{X}$; Structure $\mathbf{A}$; Iterations $I_{m}$; \\
    Number of clusters $k$; Epoch flag $l=1$; \\
    Hyper-parameter $\beta$.}
\KwOut{Clustering label $\mathbf{Y}$.} 
\begin{algorithmic}[1] 
\STATE Pre-train TIGAE with Eq. (\ref{eq:loss_TIGAE}) and initialize \SynC{};
\WHILE{$l\le I_{m}$}
\STATE Perform gradient-free forward propagation on the\\ TIGAE to predict edge probabilities $\hat{\mathbf{A}}_p^{(l)}$;
\STATE Utilize the fine-tuning strategy to enhance $\mathbf{A}_p^{(l)}$;
\STATE Perform forward propagation on the TIGAE with augmented graph to compute embeddings $\mathbf{Z}^{(l)}$;
\STATE Calculate $\mathbf{Q}$ and $\mathbf{P}$ by Eq. (\ref{eq:q}) and Eq. (\ref{eq:p});
\STATE Calculate total loss $\mathcal{L}$ by Eq. (\ref{eq:loss_total});
\STATE Perform backward propagation;
\STATE $l=l+1$;
\ENDWHILE
\STATE Obtain $\mathbf{Y}$ by Eq. (\ref{eq:y}).
\STATE \textbf{return} $\mathbf{Y}$.
\end{algorithmic}
\end{algorithm}
\section{Experimental Results and Discussion}
\label{sec:experiments}

\subsection{Experimental Setup}
\subsubsection{Datasets}
We performed extensive experiments on seven benchmark datasets (ACM, CITE, DBLP, AMAP~\cite{liu2022deep}, CORA, UAT~\cite{liu2023hard}, PubMed~\cite{kipf2016semisupervised}) to validate \SynC{}. We also tested it on Wisconsin and Texas~\cite{pei2020Ggeom} datasets with low homophily ratios and imbalanced classes. Dataset statistics are listed in Table \ref{table:settings}, where $R_h$ denotes the homophily ratio~\cite{zhu2020beyond}. Since neither our method nor the compared methods have effective large-scale training mechanisms, we did not use large-scale datasets.

\begin{table}[t]
\caption{\RevisionTwo{Statistics and experimental settings on the nine datasets.}}
\centering
\resizebox{\columnwidth}{!}{
\begin{tabular}{c|ccccc|ccc|ccc|c}
\Xhline{1pt}
\multirow{2}{*}{\textbf{Dataset}} & \multirow{2}{*}{\textbf{\#Node}} & \multirow{2}{*}{\textbf{\#Class}} & \multirow{2}{*}{\textbf{\#Dim}} & \multirow{2}{*}{\textbf{\#Edge}} & \multirow{2}{*}{$\boldsymbol{R_h}$} & \multicolumn{3}{c|}{\textbf{Pre-training}} & \multicolumn{3}{c|}{\textbf{Training}} & \multicolumn{1}{c}{\textbf{Transform}}      \\\cline{7-12}
                                  &&&&&& lr     & epoch  & $\alpha$ & lr & epoch & $\beta$& \textbf{dimension} \\ 
\Xhline{0.5pt}
\textbf{UAT}                      & 1190 & 4 & 239   & 13599 & 0.70& 2e-3 & 50 & 0 & 1e-3 &50 &1 & 128      \\
\textbf{CORA}                     & 2708 & 7 & 1433  & 5278 & 0.81& 2e-3 & 80 & 0 & 5e-3 &50 &0 & 128      \\
\textbf{ACM}                      & 3025 & 3 & 1870  & 13128 & 0.82& 1e-3 & 50 & 0 & 2e-3 &50 &1 & 512      \\
\textbf{CITE}                     & 3327 & 6 & 3703  & 4552 & 0.74& 2e-3 & 20 & 1 & 6e-3 &50 &1 & 512      \\
\textbf{DBLP}                     & 4057 & 4 & 334   & 3528 & 0.80& 2e-3 & 20 & 1 & 2e-2 &50 &1 & 512      \\
\textbf{AMAP}                     & 7650 & 8 & 745   & 119081 & 0.83& 1e-3 & 80 & 1 & 1e-4 &50 &1 & 512      \\
\textbf{\RevisionTwo{PubMed}} & \RevisionTwo{19717} & \RevisionTwo{3} & \RevisionTwo{500} & \RevisionTwo{44324} & \RevisionTwo{0.80}& \RevisionTwo{2e-3} & \RevisionTwo{50} & \RevisionTwo{10} & \RevisionTwo{1e-3} &\RevisionTwo{50} &\RevisionTwo{0} & \RevisionTwo{128} \\\midrule
\textbf{Wisconsin}                & 251 & 5 & 1703   & 515 & 0.20& 1e-2 & 20 & 1 & 1e-2 &50 &1 & 512      \\
\textbf{Texas}                    & 183 & 5 & 1703   & 325 & 0.11& 5e-3 & 20 & 10 & 5e-3 &50 &0 & 512      \\
\bottomrule
\end{tabular}
}
\label{table:settings}
\end{table}

\subsubsection{Baselines}
We compared \SynC{} with state-of-the-art methods: vanilla GAE~\cite{kipf2016variational}, SDCN~\cite{bo2020structural}, DFCN~\cite{tu2021deep}, DCRN~\cite{liu2022deep}, AGC-DRR~\cite{gong2022attributed}, CCGC~\cite{yang2023ccgc}, HSAN~\cite{liu2023hard}, R$^2$FGC~\cite{yi2023redundancy}, \RevisionOne{SCGC~\cite{liu2024simple}}, \RevisionTwo{NS4GC~\cite{liu2024reliable}}, \RevisionOne{IDCRN~\cite{liu2025improved}}, \RevisionTwo{HomoCAGC~\cite{chen2025homophily}}, \RevisionTwo{NeuCGC~\cite{peng2025trustworthy}} (detailed in related work). Partial results are cited from original papers or conducted with their open-source code. We evaluated HSAN/CCGC on ACM/DBLP and GAE/AGC-DRR on UAT via open-source codes (v0.0.1)\footnote{\href{https://github.com/Marigoldwu/PyDGC/releases/tag/v0.0.1}{A-Unified-Framework-for-Deep-Attribute-Graph-Clustering}}. NS4GC was tested via the benchmark~\cite{wu2025dgcbench}. Other models’ structures remained unchanged in comparisons. Unspecified parameters in original papers were set based on provided dataset parameters: AGC-DRR used consistent parameters across UAT and other datasets; HSAN adopted CORA parameters for ACM (no PCA on DBLP); CCGC used lr=1e-3 (DBLP) and 1e-4 (ACM) with 400 iterations (as in the paper); GAE used lr=1e-3 and 50 epochs on UAT.

\subsubsection{Metrics}
We evaluated \SynC{} using four standard graph clustering metrics: accuracy (ACC), normalized mutual information (NMI), adjusted rand index (ARI), and macro F1-score (F1), which are adopted in comparative methods’ papers~\cite{bo2020structural, liu2022deep}. Higher values indicate better performance.

\subsubsection{Training Procedure and Parameter Settings}
Table \ref{table:settings} lists detailed learning rates and hyperparameter settings. To ensure reliable clustering and initial cluster centers, we pre-trained TIGAE: the linear transformation dimension was 128 for UAT/CORA/PubMed and 512 for other datasets; $\alpha$ was set to 0 for ACM/CORA/UAT. Pre-training ran for at least 20 epochs, with pre-trained parameters loaded for subsequent 50-epoch training to ensure optimal convergence. Most experiments were conducted on a Tesla T4 GPU (16 GB) with PyTorch 2.0.1 and an NVIDIA L40 GPU (48GB); time/space consumption tests used an NVIDIA GTX 1050 GPU (2 GB) and Intel(R) Core(TM) i5-8300H CPU (2.30GHz). Ten runs were performed to report mean±std, with a fixed random seed (325) for pre-training and training to ensure reproducibility.

\begin{table*}[t]
\caption{\RevisionTwo{Clustering results on seven benchmark datasets. The best is highlighted in bold, and the asterisk (*) denotes the second best. `-’ indicates missing source code, thus unable to test. `OOM' indicates out-of-memory occurred on a single Nvidia L40 GPU (48 GB).}}
\centering
\resizebox{\textwidth}{!}{
\begin{tabular}{c|c|ccccccccccccc|cc}
\toprule
\textbf{Datasets}
&\textbf{Metrics}
&\makecell{\textbf{Vanilla GAE}\\ NeurIPS'16}
&\makecell{\textbf{SDCN}\\ WWW'20}
&\makecell{\textbf{DFCN}\\ AAAI'21}
&\makecell{\textbf{DCRN}\\ AAAI'22}
&\makecell{\textbf{AGC-DRR}\\ IJCAI'22}
&\makecell{\textbf{CCGC}\\ AAAI'23}
&\makecell{\textbf{HSAN}\\ AAAI'23}
&\makecell{\textbf{R$^2$FGC}\\ TNNLS'23}
&\makecell{\textbf{\RevisionOne{SCGC}}\\ \RevisionOne{TNNLS'24}}
&\makecell{\textbf{\RevisionTwo{NS4GC}}\\ \RevisionTwo{TKDE'24}}
&\makecell{\textbf{\RevisionOne{IDCRN}}\\ \RevisionOne{TNNLS'25}}
&\makecell{\textbf{\RevisionTwo{HomoCAGC}}\\ \RevisionTwo{TCSVT'25}}
&\makecell{\textbf{\RevisionTwo{NeuCGC}}\\ \RevisionTwo{TKDE'25}}
&\makecell{\textbf{\SynC{}}\\ KMeans}
&\makecell{\textbf{\SynC{}}\\ SSL} \\
\Xhline{0.5pt}
\multirow{4}{*}{\textbf{ACM}}
&\textbf{ACC}&84.52±1.44&90.45±0.18&90.90±0.20&91.93±0.20&{92.55±0.09}&88.76±0.91&82.50±2.47&{92.43±0.18}&\RevisionOne{89.80±0.45}&\RevisionTwo{88.82±0.73}&\RevisionOne{92.58±0.08\textsuperscript{*}}&\RevisionTwo{89.17±0.95}&\RevisionTwo{92.48±0.26}&{91.51±0.12}&{\textbf{92.73±0.04}}\\
&\textbf{NMI}&55.38±1.92&68.31±0.25&69.40±0.40&71.56±0.61&{72.89±0.24}&64.28±1.88&52.22±4.45&{72.42±0.53}&\RevisionOne{66.60±1.03}&\RevisionTwo{63.67±1.72}&\RevisionOne{73.17±0.32}&\RevisionTwo{64.96±1.80}&\RevisionTwo{\textbf{73.60±0.68}}&{70.01±0.26}&{{73.58±0.22}\textsuperscript{*}}\\
&\textbf{ARI}&59.46±3.10&73.91±0.40&74.90±0.40&77.56±0.52&{79.08±0.24}&69.86±2.09&55.12±5.58&{78.72±0.47}&\RevisionOne{72.37±1.09}&\RevisionTwo{69.76±1.77}&\RevisionOne{79.18±0.22\textsuperscript{*}}&\RevisionTwo{70.62±2.26}&\RevisionTwo{79.02±0.67}&{76.43±0.30}&{\textbf{79.58±0.11}}\\
&\textbf{F1} &84.65±1.33&90.42±0.19&90.80±0.20&91.94±0.20&{92.55±0.09}&88.71±0.91&82.43±2.51&{92.45±0.18}&\RevisionOne{89.74±0.45}&\RevisionTwo{88.81±0.73}&\RevisionOne{92.60±0.08\textsuperscript{*}}&\RevisionTwo{89.17±0.93}&\RevisionTwo{92.47±0.26}&{91.49±0.12}&{\textbf{92.74±0.04}}\\ \midrule
\multirow{4}{*}{\textbf{DBLP}}
&\textbf{ACC}&61.21±1.22&68.05±1.81&76.00±0.80&79.66±0.25&80.41±0.47&65.19±4.20&71.91±0.96&{80.95±0.20}&\RevisionOne{67.25±1.64}&\RevisionTwo{79.19±0.42}&\RevisionOne{82.08±0.18\textsuperscript{*}}&\RevisionTwo{66.49±1.34}&\RevisionTwo{81.31±0.12}&{77.62±0.55}&{\textbf{83.48±0.13}}\\
&\textbf{NMI}&30.80±0.91&39.50±1.34&43.70±1.00&48.95±0.44&49.77±0.65&33.48±3.94&42.74±0.63&{50.82±0.32}&\RevisionOne{38.64±1.25}&\RevisionTwo{48.23±0.71}&\RevisionOne{52.70±0.36\textsuperscript{*}}&\RevisionTwo{36.31±1.50}&\RevisionTwo{52.55±0.27}&{45.62±1.54}&{\textbf{55.11±0.24}}\\
&\textbf{ARI}&22.02±1.40&39.15±2.01&47.00±1.50&53.60±0.46&55.39±0.88&34.21±5.43&43.81±1.29&{56.34±0.42}&\RevisionOne{35.95±1.46}&\RevisionTwo{53.14±0.86}&\RevisionOne{58.81±0.37\textsuperscript{*}}&\RevisionTwo{28.36±1.88}&\RevisionTwo{58.12±0.24}&{50.17±1.18}&{\textbf{61.70±0.27}}\\
&\textbf{F1} &61.41±2.23&67.71±1.51&75.70±0.80&79.28±0.26&79.90±0.45&64.00±4.29&71.13±1.23&{80.54±0.19}&\RevisionOne{67.06±1.73}&\RevisionTwo{78.77±0.40}&\RevisionOne{81.47±0.20\textsuperscript{*}}&\RevisionTwo{66.54±1.77}&\RevisionTwo{80.71±0.12}&{77.19±0.54}&{\textbf{82.90±0.17}}\\ \midrule
\multirow{4}{*}{\textbf{CITE}}
&\textbf{ACC}&61.35±0.80&65.96±0.31&69.50±0.20&70.86±0.18&68.32±1.83&69.84±0.94&71.15±0.80&{70.60±0.45}&\RevisionOne{71.02±0.77}&\RevisionTwo{68.42±0.51}&\RevisionOne{71.40±0.08}&\RevisionTwo{68.90±1.07}&\RevisionTwo{\textbf{72.75±0.50}}&{66.93±0.80}&{{71.77±0.27}\textsuperscript{*}}\\
&\textbf{NMI}&34.63±0.65&38.71±0.32&43.90±0.20&{45.86±0.35}&43.28±1.41&44.33±0.79&45.06±0.74&{45.39±0.37}&\RevisionOne{45.25±0.45}&\RevisionTwo{43.51±0.53}&\RevisionOne{{46.77±0.21}\textsuperscript{*}}&\RevisionTwo{44.13±0.92}&\RevisionTwo{\textbf{47.00±0.52}}&{40.49±0.53}&{46.37±0.42}\\
&\textbf{ARI}&33.55±1.18&40.17±0.43&45.50±0.30&{47.64±0.30}&45.34±2.33&45.68±1.80&47.05±1.12&{47.07±0.30}&\RevisionOne{46.29±1.13}&\RevisionTwo{44.54±0.66}&\RevisionOne{\textbf{48.67±0.20}}&\RevisionTwo{44.63±1.65}&\RevisionTwo{48.58±1.02\textsuperscript{*}}&{41.52±0.88}&{{48.09±0.45}}\\
&\textbf{F1} &57.36±0.82&63.62±0.24&64.30±0.20&{65.83±0.21\textsuperscript{*}}&64.82±1.60&62.71±2.06&63.01±1.79&{65.28±0.12}&\RevisionOne{64.80±1.01}&\RevisionTwo{63.99±0.48}&\RevisionOne{\textbf{66.27±0.21}}&\RevisionTwo{64.75±0.51}&\RevisionTwo{64.26±0.72}&{62.97±0.36}&{65.72±0.36}\\ \midrule
\multirow{4}{*}{{\textbf{CORA}}}  
&{\textbf{ACC}}&{68.09±2.59}&{35.60±2.83}&{36.33±0.49}&{61.93±0.47}&{40.62±0.55}&{73.88±1.20}&{77.07±1.56}&{-}&\RevisionOne{73.88±0.88}&\RevisionTwo{75.79±0.63}&\RevisionOne{-}&\RevisionTwo{73.82±2.86}&\RevisionTwo{76.89±0.88}&{77.22±0.40\textsuperscript{*}}&{\textbf{78.58±0.38}} \\
&{\textbf{NMI}}&{47.77±2.53}&{14.28±1.91}&{19.36±0.87}&{45.13±1.57}&{18.74±0.73}&{56.45±1.04}&{\textbf{59.21±1.03}}&{-}&\RevisionOne{56.10±0.72}&\RevisionTwo{59.77±0.64}&\RevisionOne{-}&\RevisionTwo{57.41±1.41}&\RevisionTwo{58.56±0.76}&{57.26±0.57}&{58.13±0.52\textsuperscript{*}} \\
&{\textbf{ARI}}&{42.52±3.10}&{07.78±3.24}&{04.67±2.10}&{33.15±0.14}&{14.80±1.64}&{52.51±1.89}&{57.52±2.70\textsuperscript{*}}&{-}&\RevisionOne{51.79±1.59}&\RevisionTwo{56.99±1.28}&\RevisionOne{-}&\RevisionTwo{54.42±3.13}&\RevisionTwo{55.85±1.57}&{55.34±0.64}&{\textbf{57.90±1.06}} \\
&{\textbf{F1} }&{67.11±2.71}&{24.37±1.04}&{26.16±0.50}&{49.50±0.42}&{31.23±0.57}&{70.98±2.79}&{75.11±1.40}&{-}&\RevisionOne{70.81±1.96}&\RevisionTwo{74.41±0.58}&\RevisionOne{-}&\RevisionTwo{70.83±4.01}&\RevisionTwo{75.46±1.04}&{76.31±0.41\textsuperscript{*}}&{\textbf{77.65±0.30}} \\ \midrule
\multirow{4}{*}{\textbf{AMAP}}
&\textbf{ACC}&71.57±2.48&53.44±0.81&76.88±0.80&79.94±0.13&78.11±1.69&77.25±0.41&77.02±0.33&{81.28±0.05\textsuperscript{*}}&\RevisionOne{77.48±0.37}&\RevisionTwo{78.06±1.90}&\RevisionOne{80.17±0.04}&\RevisionTwo{70.78±3.26}&\RevisionTwo{79.39±0.28}&{80.94±0.11}&{\textbf{82.48±0.04}}\\
&\textbf{NMI}&62.13±2.79&44.85±0.83&69.21±1.00&73.70±0.24&72.21±1.63&67.44±0.48&67.21±0.33&{{73.88±0.17}\textsuperscript{*}}&\RevisionOne{67.67±0.88}&\RevisionTwo{65.63±2.38}&\RevisionOne{\textbf{74.32±0.07}}&\RevisionTwo{62.32±2.39}&\RevisionTwo{72.67±0.49}&{68.92±0.18}&{69.70±0.23}\\
&\textbf{ARI}&48.82±4.57&31.21±1.23&58.98±0.84&63.69±0.20&61.15±1.65&57.99±0.66&58.01±0.48&{\textbf{66.25±0.36}}&\RevisionOne{58.48±0.72}&\RevisionTwo{57.43±2.98}&\RevisionOne{64.10±0.10}&\RevisionTwo{48.28±4.00}&\RevisionTwo{62.83±0.37}&{61.85±0.54}&{65.02±0.11\textsuperscript{*}}\\
&\textbf{F1} &68.08±1.76&50.66±1.49&71.58±0.31&73.82±0.12&72.72±0.97&72.18±0.57&72.03±0.46&{75.29±0.32}&\RevisionOne{72.22±0.97}&\RevisionTwo{76.52±2.13}&\RevisionOne{74.01±0.04}&\RevisionTwo{73.45±0.45}&\RevisionTwo{67.22±2.90}&{79.51±0.40\textsuperscript{*}}&{\textbf{80.69±0.11}}\\ \midrule
\multirow{4}{*}{\textbf{UAT}}
&\textbf{ACC}&55.18±0.46&52.25±1.91&33.61±0.09&49.92±1.25&52.57±1.44&56.34±1.11&56.04±0.67&{-}&\RevisionOne{56.58±1.62}&\RevisionTwo{55.25±0.46}&\RevisionOne{-}&\RevisionTwo{55.18±0.93}&\RevisionTwo{57.46±0.62\textsuperscript{*}}&{\textbf{60.39±0.46}}&{57.33±0.13}\\
&\textbf{NMI}&23.40±0.96&21.61±1.26&26.49±0.41&24.09±0.53&23.75±1.47&28.15±1.92&26.99±2.11&{-}&\RevisionOne{28.07±0.71}&\RevisionTwo{22.31±0.51}&\RevisionOne{-}&\RevisionTwo{26.62±1.31}&\RevisionTwo{28.16±1.57}&{\textbf{30.26±0.62}}&{28.58±0.24\textsuperscript{*}}\\
&\textbf{ARI}&22.44±1.34&21.63±1.49&11.87±0.23&17.17±0.69&21.73±0.85&25.52±2.09&25.22±1.96&{-}&\RevisionOne{24.80±1.85}&\RevisionTwo{22.32±0.60}&\RevisionOne{-}&\RevisionTwo{20.40±2.19}&\RevisionTwo{26.82±0.80\textsuperscript{*}}&{\textbf{30.09±0.76}}&{26.60±0.17}\\
&\textbf{F1} &54.39±0.89&45.59±3.54&25.79±0.29&44.81±0.87&51.97±2.10&55.24±1.69&54.20±1.84&{-}&\RevisionOne{55.52±0.87}&\RevisionTwo{51.76±0.52}&\RevisionOne{-}&\RevisionTwo{55.31±0.87}&\RevisionTwo{55.35±1.26}&{\textbf{58.91±1.02}}&{57.34±0.23\textsuperscript{*}}\\ \midrule
\multirow{4}{*}{\RevisionTwo{\textbf{PubMed}}}
&\RevisionTwo{\textbf{ACC}}&\RevisionTwo{62.09±0.81}&\RevisionTwo{64.20±1.30}&\RevisionTwo{68.89±0.07}&\RevisionTwo{69.87±0.07}&\RevisionTwo{64.00±6.85}&\RevisionTwo{47.74±2.14}&\RevisionTwo{OOM}&\RevisionTwo{-}&\RevisionTwo{47.57±2.39}&\RevisionTwo{70.56±0.08}&\RevisionTwo{70.02±0.03}&\RevisionTwo{OOM}&\RevisionTwo{65.14±0.17}&\RevisionTwo{70.91±0.70\textsuperscript{*}}&\RevisionTwo{\textbf{71.96±0.08}}\\
&\RevisionTwo{\textbf{NMI}}&\RevisionTwo{23.84±3.54}&\RevisionTwo{22.87±2.04}&\RevisionTwo{31.43±0.13}&\RevisionTwo{32.20±0.08}&\RevisionTwo{23.72±4.24}&\RevisionTwo{07.44±1.72}&\RevisionTwo{OOM}&\RevisionTwo{-}&\RevisionTwo{07.34±1.97}&\RevisionTwo{32.84±0.23\textsuperscript{*}}&\RevisionTwo{\textbf{33.29±0.07}}&\RevisionTwo{OOM}&\RevisionTwo{25.42±0.34}&\RevisionTwo{31.87±0.47}&\RevisionTwo{31.60±0.17}\\
&\RevisionTwo{\textbf{ARI}}&\RevisionTwo{20.62±1.39}&\RevisionTwo{22.30±2.07}&\RevisionTwo{30.64±0.11}&\RevisionTwo{31.41±0.12}&\RevisionTwo{24.87±4.86}&\RevisionTwo{05.55±2.42}&\RevisionTwo{OOM}&\RevisionTwo{-}&\RevisionTwo{05.40±2.46}&\RevisionTwo{33.48±0.13}&\RevisionTwo{32.67±0.05}&\RevisionTwo{OOM}&\RevisionTwo{25.42±0.28}&\RevisionTwo{33.63±0.57\textsuperscript{*}}&\RevisionTwo{\textbf{34.66±0.15}}\\
&\RevisionTwo{\textbf{F1}} &\RevisionTwo{61.37±0.85}&\RevisionTwo{65.01±1.21}&\RevisionTwo{68.10±0.07}&\RevisionTwo{68.94±0.08}&\RevisionTwo{64.42±4.59}&\RevisionTwo{45.56±2.23}&\RevisionTwo{OOM}&\RevisionTwo{-}&\RevisionTwo{45.37±2.37}&\RevisionTwo{69.56±0.13}&\RevisionTwo{69.19±0.03}&\RevisionTwo{OOM}&\RevisionTwo{65.51±0.17}&\RevisionTwo{70.32±0.67\textsuperscript{*}}&\RevisionTwo{\textbf{71.45±0.07}}\\ \midrule
\midrule
\multicolumn{2}{c|}{\RevisionTwo{\textbf{Average Rank}}} &\RevisionTwo{15}	&\RevisionTwo{14}	&\RevisionTwo{11}	&\RevisionTwo{5}	&\RevisionTwo{6}	&\RevisionTwo{12}	&\RevisionTwo{10}	&\RevisionTwo{8}	&\RevisionTwo{9}	&\RevisionTwo{7}	&\RevisionTwo{4}	&\RevisionTwo{13}	&\RevisionTwo{2\textsuperscript{*}}	&\RevisionTwo{3}	&\RevisionTwo{\textbf{1}} \\
\bottomrule
\end{tabular}
}
\label{table:comparison}
\end{table*}

\subsection{Comparative analysis}
Table \ref{table:comparison} shows the clustering results regarding various baselines and metrics. We make the following key insights:
\begin{enumerate}
    \item Compared with classical deep clustering methods, \SynC{} improves input data quality from both attribute and structural aspects, fully utilizing GCNs’ strong aggregation ability. \RevisionOne{Notably, with the same GAE backbone, it outperforms GAE on DBLP by 22\% in ACC and 39.67\% in ARI. Against state-of-the-art methods, it also achieves 2.41\% higher NMI and 2.89\% higher ARI on DBLP.}
    \item \SynC{} still delivers competitive performance on many datasets, even with K-Means alone, notably on the small-scale UAT dataset, validating the effectiveness of our proposed embedding learning approach.
    \item {Methods (e.g., AGC-DRR, R$^2$FGC) that employ specific attribute or structure learning and optimization strategies tend to outperform those that neglect such considerations}, which is because GCN is unable to detect and process anomalous information during the learning process effectively, and unprocessed data may impede the propagation of meaningful information, thereby compromising the discriminability of the learned embeddings.
    \item While contrastive learning-based methods (e.g., CCGC, HSAN, \RevisionOne{SCGC}) have shown favorable clustering results, the significant variations of standard deviation indicate that they may lack stability, making it impossible to apply them practically without supervision.
    \item \RevisionTwo{Recent DGC methods based on contrastive learning mostly focus on constructing positive and negative sample pairs and learning node semantic information (e.g., NS4GC, HomoCAGC, NeuCGC). However, due to failing to consider the reciprocal relationship between structure learning and feature learning, their performance on most datasets is inferior to our \SynC{}.}
    \item {We also note that the performance of our \SynC{} on NMI and ARI improves slightly on the class-imbalanced datasets (e.g., AMAP) because the embeddings learned by GCN will feature the dominant classes better.} 
\end{enumerate}

In conclusion, the significant improvements demonstrate that our proposed \SynC{} exhibits excellent clustering performance on common benchmark datasets.

\subsection{Ablation Study}
\subsubsection{Modules Analysis}
\begin{figure}[t]
  \centering
  
  \subfigure[ACM]{
    \includegraphics[trim=0 0 30pt 30pt, clip, width=0.45\columnwidth]{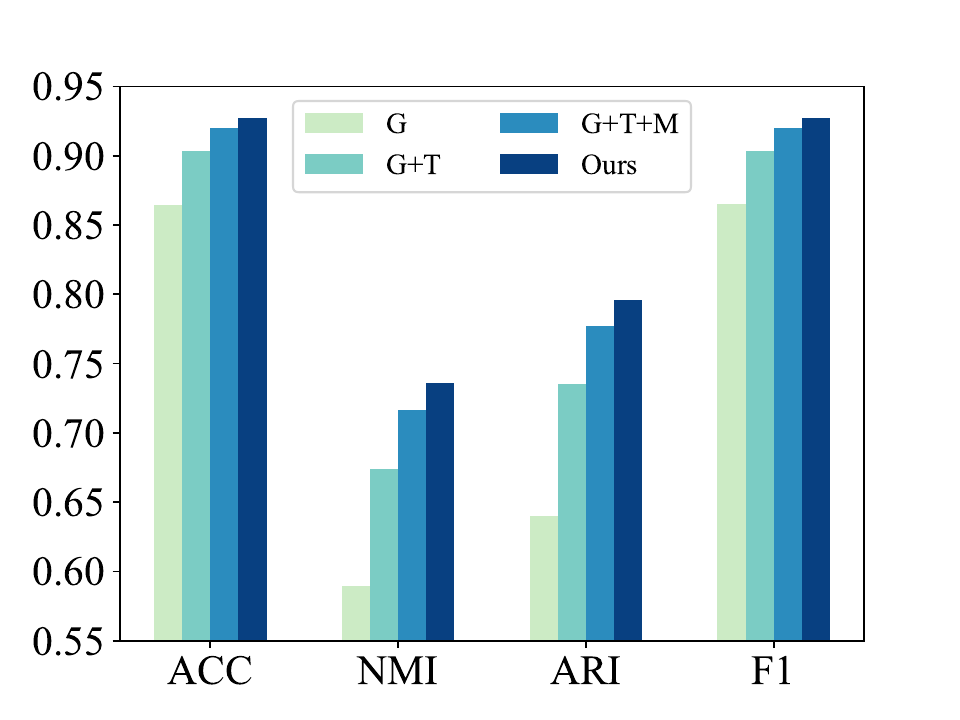}
    \label{fig:ablation_acm}
  }
  \hfill
  \subfigure[CITE]{
    \includegraphics[trim=0 0 30pt 30pt, clip, width=0.45\columnwidth]{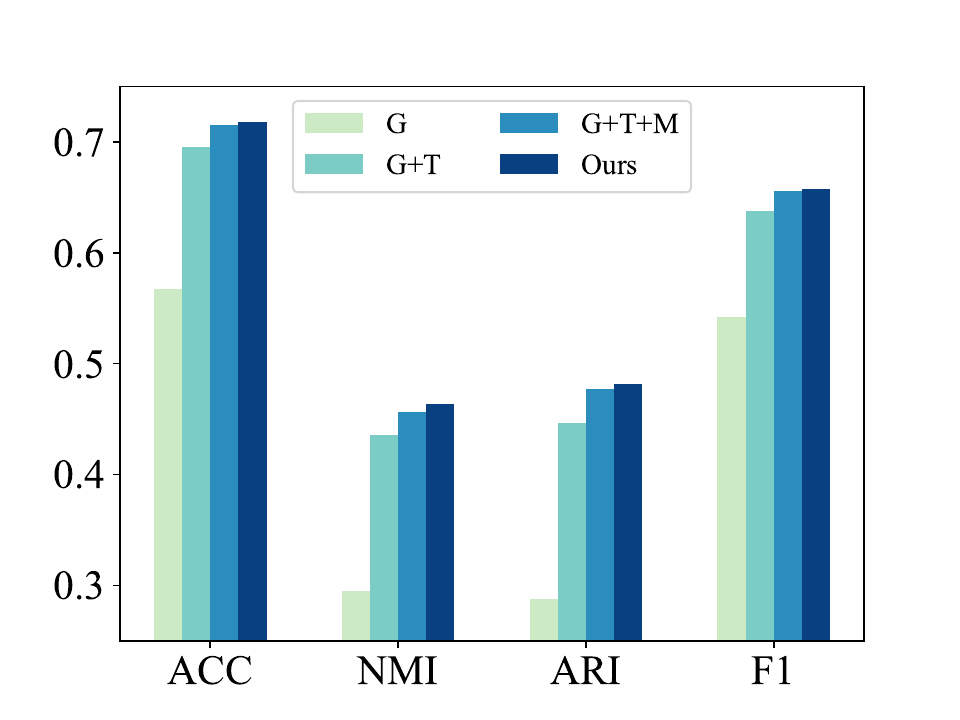}
    \label{fig:ablation_cite}
  }

  \subfigure[DBLP]{
    \includegraphics[trim=0 0 30pt 30pt, clip, width=0.45\columnwidth]{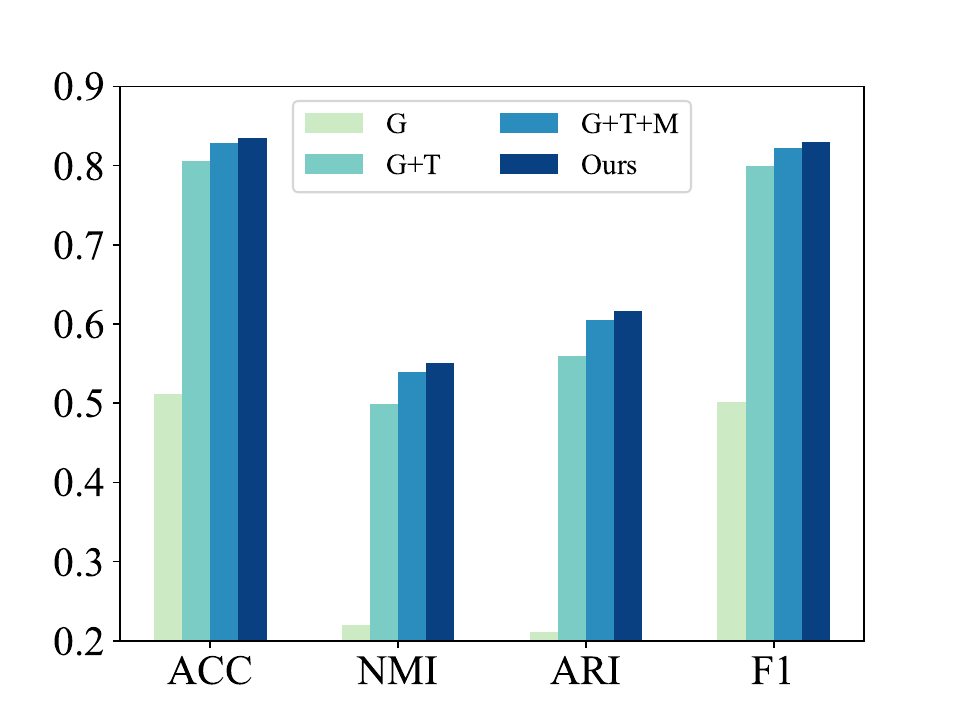}
    \label{fig:ablation_dblp}
  }
  \hfill
  \subfigure[UAT]{
    \includegraphics[trim=0 0 30pt 30pt, clip, width=0.45\columnwidth]{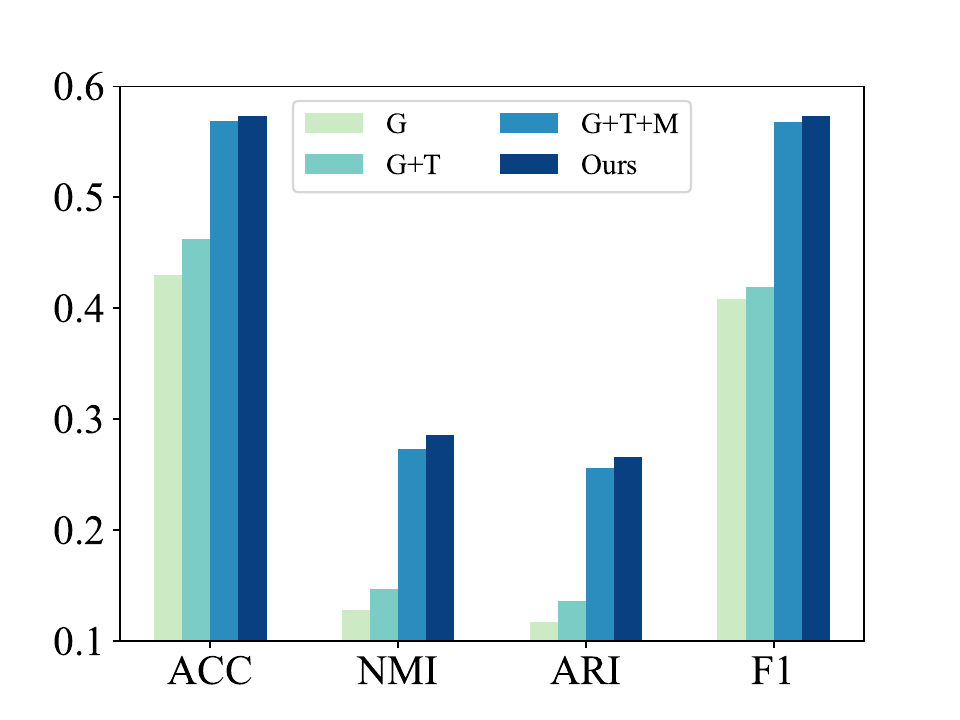}
    \label{fig:ablation_uat}
  }

  \caption{Ablation results of the proposed \SynC{} on four datasets.}
  \label{fig:ablation}
\end{figure}

We designed three variants and performed ablation experiments to verify the effectiveness of each module in our \SynC{}. Variant details are listed below, with results shown in Fig.~\ref{fig:ablation}.

\begin{itemize}
    \item ``G": Vanilla GAE, used as the baseline.
    \item ``G+T": Our proposed TIGAE; comparing with ``G" shows its effectiveness in addressing GAE's representation collapse.
    \item ``G+T+M": Our synergistic interaction framework based on TIGAE; comparing with ``G+T" validates the advantages of synergistic representation learning and structure augmentation, while comparing with \SynC{} reveals the effect of structure fine-tuning.
\end{itemize}

As shown in Fig.~\ref{fig:ablation}, all modules in our proposed \SynC{} enhance clustering performance. Specifically, ``G+T" outperforms ``G" because TIGAE integrates explicit structural information. Moreover, comparing ``G+T" with ``G+T+M" confirms that the TIGAE-based synergistic interaction framework boosts performance, particularly on UAT. Though structure fine-tuning (comparing \SynC{} with ``G+T+M") does not significantly enhance performance on high-homophily datasets, it prevents the predicted graph from deviating from the original. Overall, the TIGAE-based synergistic DGC network effectively improves clustering performance.

\begin{table}[t]
\caption{Clustering with different structure fine-tuning factors.}
\centering
\begin{tabular}{c|ccc}
\toprule
Weighting &       &           &$\surd$ \\
Link &             &$\surd$  &$\surd$ \\
Pruning  &$\surd$  &$\surd$  &$\surd$ \\
\midrule
ACM & 91.99±0.06 & 92.55±0.13 & \textbf{92.73±0.05}\\
DBLP & 82.86±0.09&83.30±0.26&\textbf{83.47±0.15}\\
CORA &77.77±0.44 &78.09±0.47 &\textbf{78.58±0.38} \\
Wisconsin & 55.34±1.25 &58.13±0.88 &\textbf{59.64±1.02} \\
Texas & 60.49±1.34 & 62.73±1.06 & \textbf{64.37±0.80} \\

\bottomrule
\end{tabular}
\label{table:ablation_sf}
\end{table}

\begin{table}[t]
\caption{{Comparison of clustering results from different methods on the Wisconsin and Texas dataset.}}
\centering
\resizebox{\columnwidth}{!}{
\begin{tabular}{c|c|cccc}
\toprule
{\textbf{Dataset}}&\textbf{Methods}&ACC &NMI &ARI &F1 \\\midrule
\multirow{6}{*}{\rotatebox{90}{{Wisconsin}}}
&\textbf{SDCN}&50.48±1.72&9.47±2.80&8.56±3.83&23.23±3.49\\
&\textbf{DFCN}&50.72±0.56&11.48±0.48&12.92±0.56&25.26±0.90\\
&\textbf{DCRN}&51.20±0.32&8.89±0.57&7.89±0.69&23.88±0.47\\ 
&\textbf{HSAN}&47.97±0.41&10.31±0.78&12.04±0.69&28.84±0.62\\ \cline{2-6}
&\textbf{w/o SF}&54.06±1.32\textsuperscript{*}&24.61±2.87\textsuperscript{*}&19.01±1.98\textsuperscript{*}&33.39±1.96\textsuperscript{*} \\
&\textbf{\SynC{}}&\textbf{59.64±1.02}&\textbf{32.79±1.42}&\textbf{26.86±1.30}&\textbf{38.19±1.30} \\ \midrule
\multirow{6}{*}{\rotatebox{90}{{Texas}}}
&\textbf{SDCN}&59.40±1.71&12.10±3.30&14.19±5.66&26.60±3.85\\
&\textbf{DFCN}&60.60±0.79&14.57±0.62&26.71±1.47\textsuperscript{*}&30.71±1.03\\
&\textbf{DCRN}&61.26±0.45&16.00±2.02&17.87±2.72&29.06±1.92\\ 
&\textbf{HSAN}&59.78±0.36&14.64±0.36&24.45±0.48&35.29±1.50\textsuperscript{*}\\ \cline{2-6}
&\textbf{w/o SF}&59.02±1.42\textsuperscript{*}&21.53±1.66\textsuperscript{*}&22.97±2.14&34.19±2.46 \\
&\textbf{\SynC{}}&\textbf{64.37±0.80}&\textbf{27.61±1.00}&\textbf{32.65±1.91}&\textbf{39.49±1.53} \\
\bottomrule
\end{tabular}
}
\label{table:lowrh_texas_wisc}
\end{table}

\begin{table}[!t]
\centering
\caption{\RevisionOne{Clustering on ACM and DBLP with different similarity measurements in the structure fine-tuning strategy.}}
\label{tab:similarity}
\resizebox{\columnwidth}{!}{
\begin{tabular}{cccccc} \toprule

\textbf{Dataset}               & \textbf{Metric}   & \textbf{Cosine}     & \textbf{Euclidean} & \textbf{Manhatten} & \textbf{Pearson}    \\ \midrule
\multirow{5}{*}{\textbf{ACM}}  & \textbf{ACC}      & \textbf{92.73±0.04} & 90.40±0.44         & 91.50±0.07         & { 91.52±0.09}\textsuperscript{*}    \\
                               & \textbf{NMI}      & \textbf{73.58±0.22} & 67.49±0.95         & 70.41±0.20         & { 70.47±0.27}\textsuperscript{*}    \\
                               & \textbf{ARI}      & \textbf{79.58±0.11} & 73.68±1.05         & 76.47±0.18         & { 76.51±0.24}\textsuperscript{*}    \\
                               & \textbf{F1}       & \textbf{92.74±0.04} & 90.37±0.45         & 91.49±0.07         & { 91.51±0.10}\textsuperscript{*}    \\
                               & \textbf{Time (s)} & \textbf{1.1}        & 3.5\textsuperscript{*}                & 7.4                & \textbf{1.1}        \\ \midrule
\multirow{5}{*}{\textbf{DBLP}} & \textbf{ACC}      & { 83.48±0.13}\textsuperscript{*}    & 82.56±0.13         & 82.47±0.16         & \textbf{83.53±0.20} \\
                               & \textbf{NMI}      & { 55.11±0.24}\textsuperscript{*}    & 53.57±0.23         & 53.34±0.31         & \textbf{55.20±0.35} \\
                               & \textbf{ARI}      & { 61.70±0.27}\textsuperscript{*}    & 59.75±0.23         & 59.57±0.31         & \textbf{61.82±0.41} \\
                               & \textbf{F1}       & { 82.90±0.17}\textsuperscript{*}    & 82.03±0.15         & 81.93±0.19         & \textbf{82.95±0.22} \\
                               & \textbf{Time (s)} & \textbf{1.8}        & { 1.9}\textsuperscript{*}          & 13.2               & { 1.9}\textsuperscript{*}          \\ \bottomrule
\end{tabular}}
\end{table}

\subsubsection{Structure Fine-tuning Strategy Analysis}
While the Structure Fine-tuning (SF) strategy has higher complexity and yields only marginal gains on high-homophily graphs, its effectiveness is notably improved on low-homophily and class-imbalanced datasets. Table~\ref{table:ablation_sf} reports ablation results for key components of SF, verifying that `Pruning', `Link', and `Weighting' are all effective. Notably, `Link' delivers the most substantial clustering improvements on the Wisconsin and Texas datasets.

We further compared our method with state-of-the-art approaches on the Wisconsin and Texas datasets, with results shown in Table~\ref{table:lowrh_texas_wisc}. Our method achieves remarkable clustering performance gains, primarily attributed to the SF strategy. Notably, even without SF (w/o SF), our method still delivers competitive results relative to baselines.

\subsubsection{Similarity Metric Analysis}
\label{sec:similarity}
\RevisionOne{For the similarity computation in Eq. (\ref{eq:sx}), we performed ablation experiments using four metrics: cosine similarity, Euclidean distance, Manhattan distance, and Pearson correlation coefficient. Table \ref{tab:similarity} shows that cosine similarity and Pearson correlation coefficient yield better performance, while the other metrics also perform adequately. Overall, the choice of similarity metric in Eq. (\ref{eq:sx}) has no significant impact on subsequent clustering results, enabling flexible selection according to data characteristics.}

\subsection{Hyper-parameter Analysis}
\begin{figure}[t]
    \centering
        \subfigure[Pre-training with different $\alpha$.]{
            \includegraphics[trim=40pt 0pt 40pt 20pt,clip,width=0.45\columnwidth]{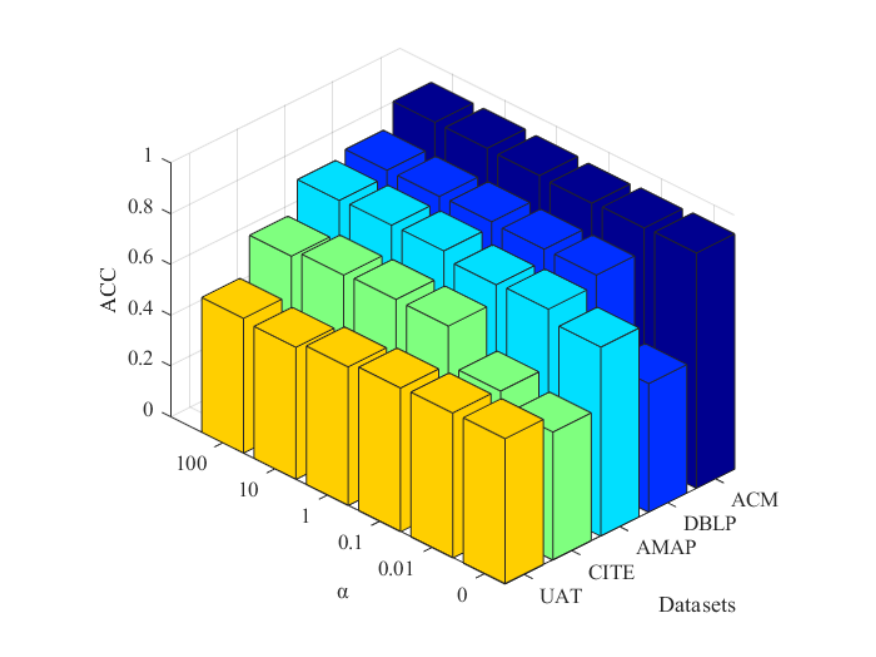}
            \label{fig:alpha}
        }
    \hfill
        \subfigure[Training with different $\beta$.]{
            \includegraphics[trim=40pt 0pt 40pt 20pt,clip,width=0.45\columnwidth]{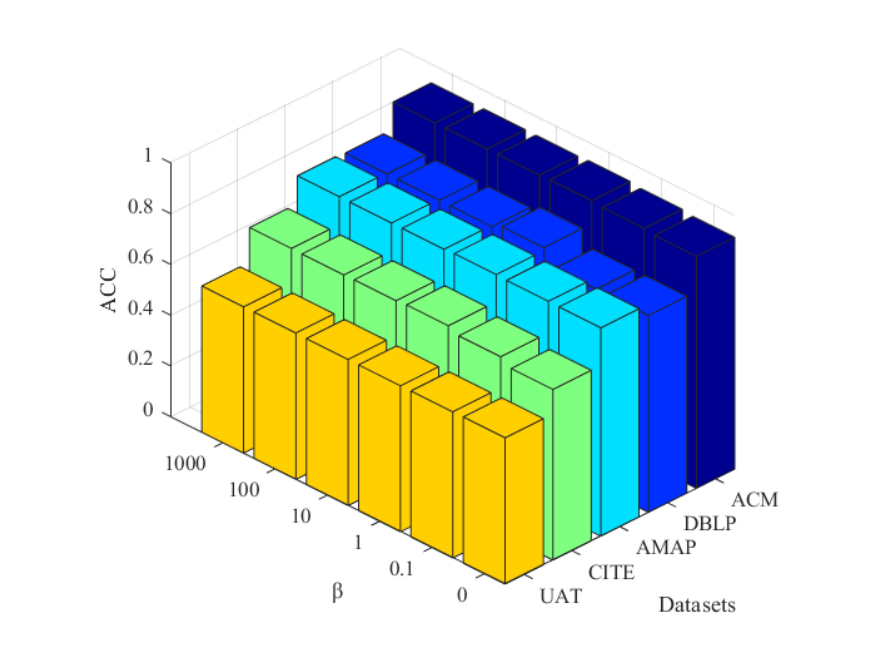}
            \label{fig:beta}
        }
    \caption{Clustering results with different hyper-parameters.}
\end{figure}

Regarding the hyper-parameter $\alpha$ in Eq. (\ref{eq:loss_TIGAE}), we conducted sensitivity experiments by selecting $\alpha$ from \{0, 0.01, 0.1, 1, 10, 100\} and selecting $\beta$ from \{0, 0.1, 1, 10, 100, 1000\}. As shown in Fig. \ref{fig:alpha}, $\alpha$ does not significantly impact the results for ACM, AMAP, and UAT. The function of $\alpha$ is to control how much the original data similarity is preserved during the linear transformation. Therefore, the tiny similarity differentiation makes $\alpha$ less meaningful for these datasets. However, for the other datasets, feature similarity of raw data provides crucial information for preventing arbitrary linear transformation. Fortunately, $\alpha$ is insensitive to values other than 0 for these datasets. Fig. \ref{fig:beta} shows that our method is insensitive to $\beta$, which is usually set to 1. Nevertheless, optimizing using the cross-entropy loss between the auxiliary distribution $\mathbf{P}$ and the clustering distribution $\mathbf{Q}$ is not always effective. For example, the best results are achieved on the CORA and Texas datasets when $\beta$ is set to 0. Even so, we still believe that \SynC{} is not sensitive to the value of $\beta$. Consequently, \SynC{} is only sensitive to whether to set $\alpha$ to 0 when the feature similarity is informative.

\begin{figure*}[!t]
  \centering
\subfigure[GAE]{
  \begin{minipage}{0.1\textwidth}
    \centerline{\includegraphics[trim=30pt 30pt 30pt 30pt, clip, width=\linewidth]{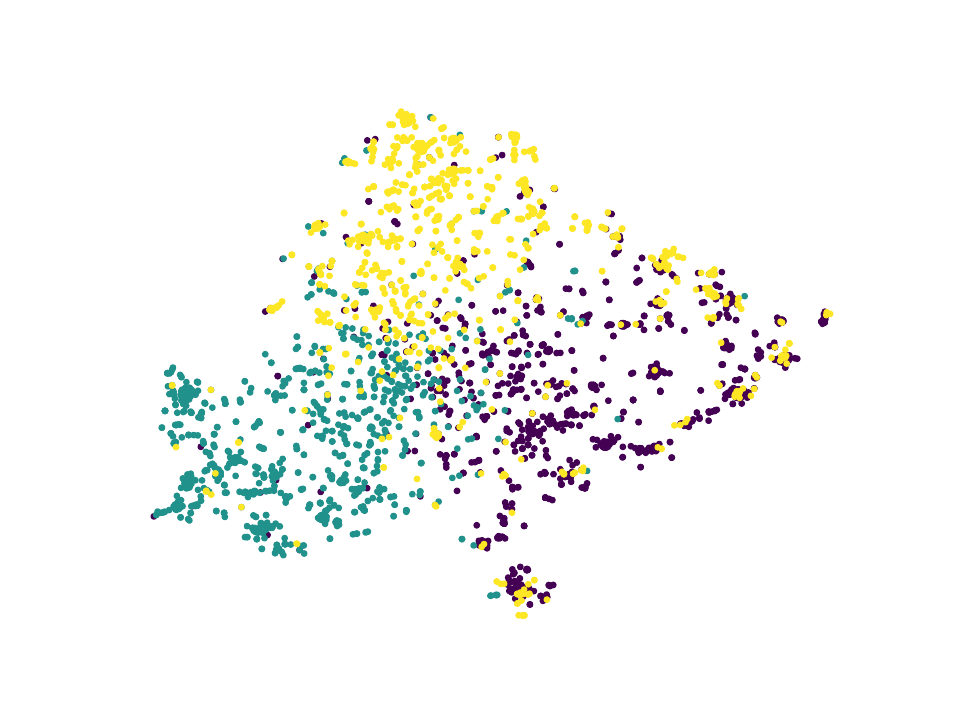}}
    \centerline{\includegraphics[trim=30pt 30pt 30pt 30pt, clip, width=\linewidth]{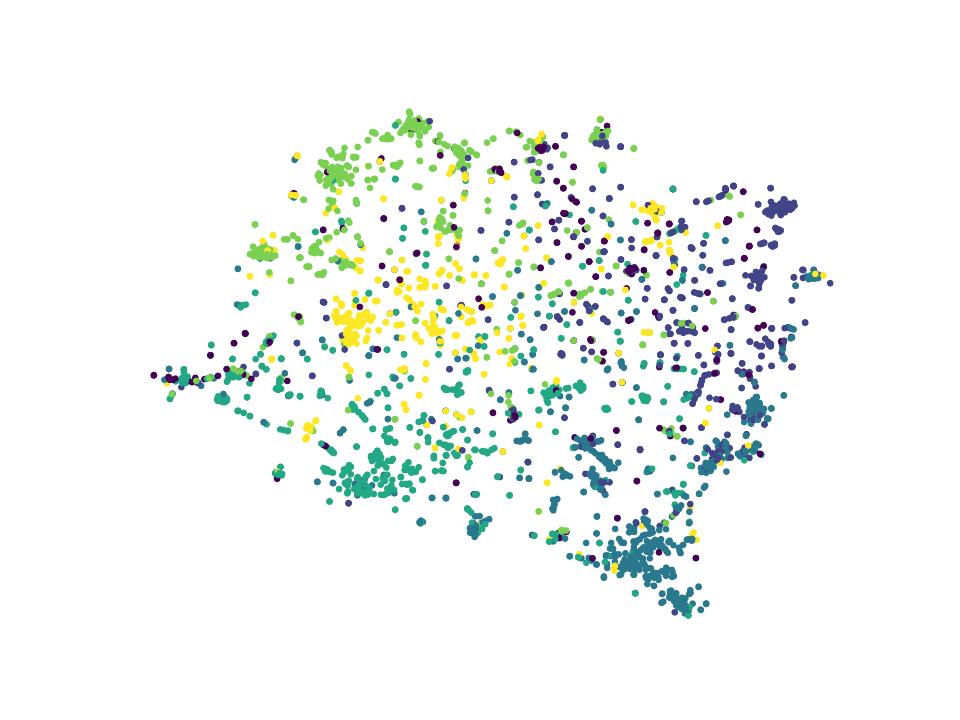}}
    \centerline{\includegraphics[trim=30pt 30pt 30pt 30pt, clip, width=\linewidth]{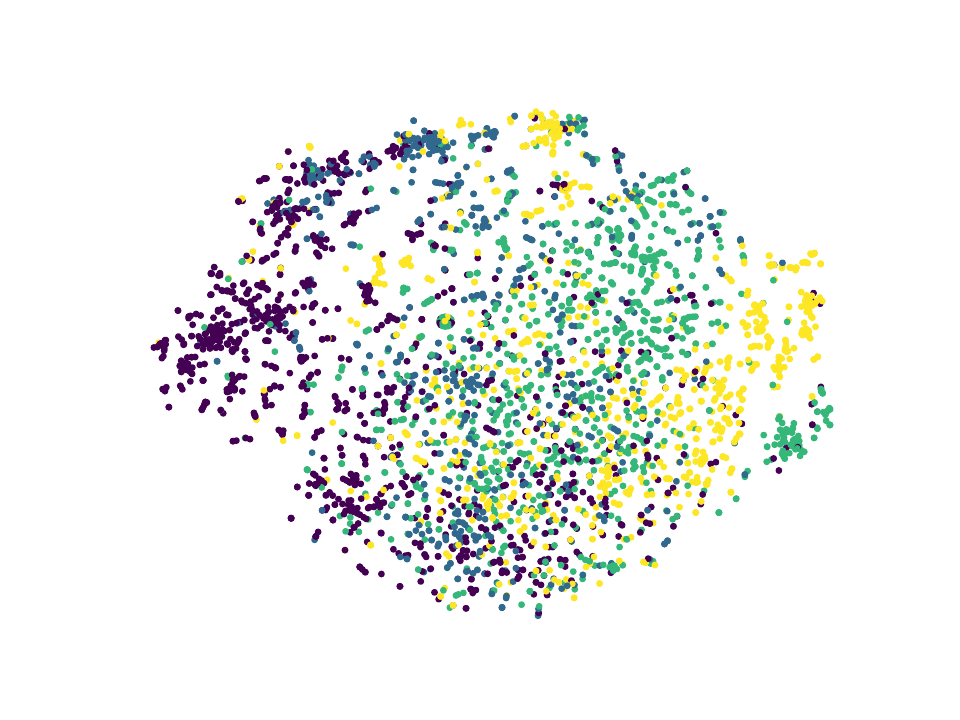}}
  \end{minipage}
\label{fig:visual_gae}
}
\hfill
\subfigure[SDCN]{
  \begin{minipage}{0.1\textwidth}
    \centerline{\includegraphics[trim=30pt 30pt 30pt 30pt, clip, width=\linewidth]{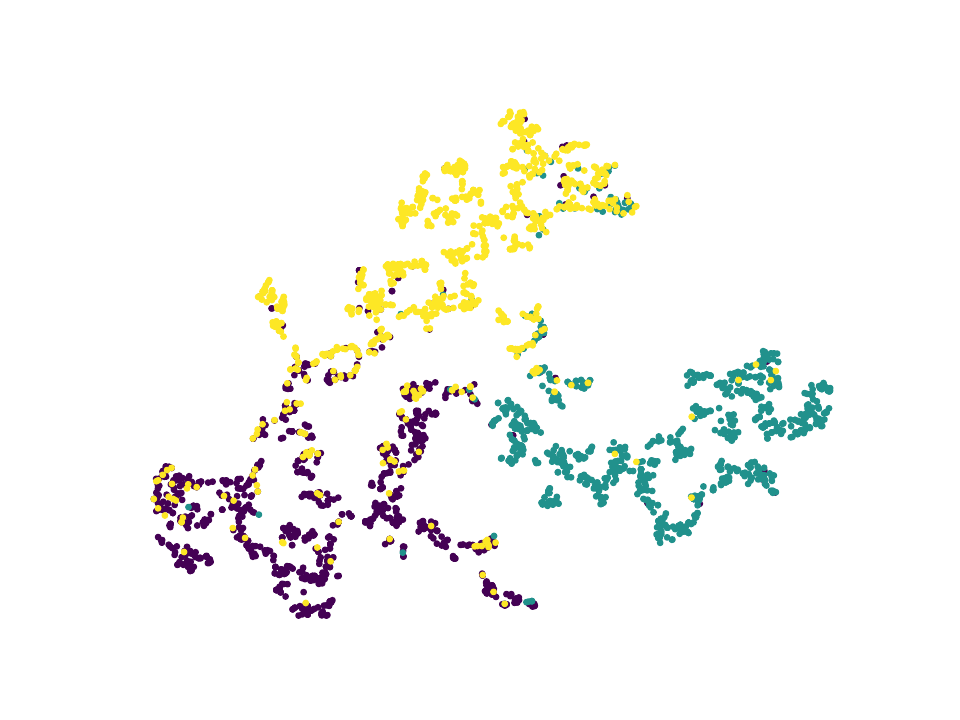}}
    \centerline{\includegraphics[trim=30pt 30pt 30pt 30pt, clip, width=\linewidth]{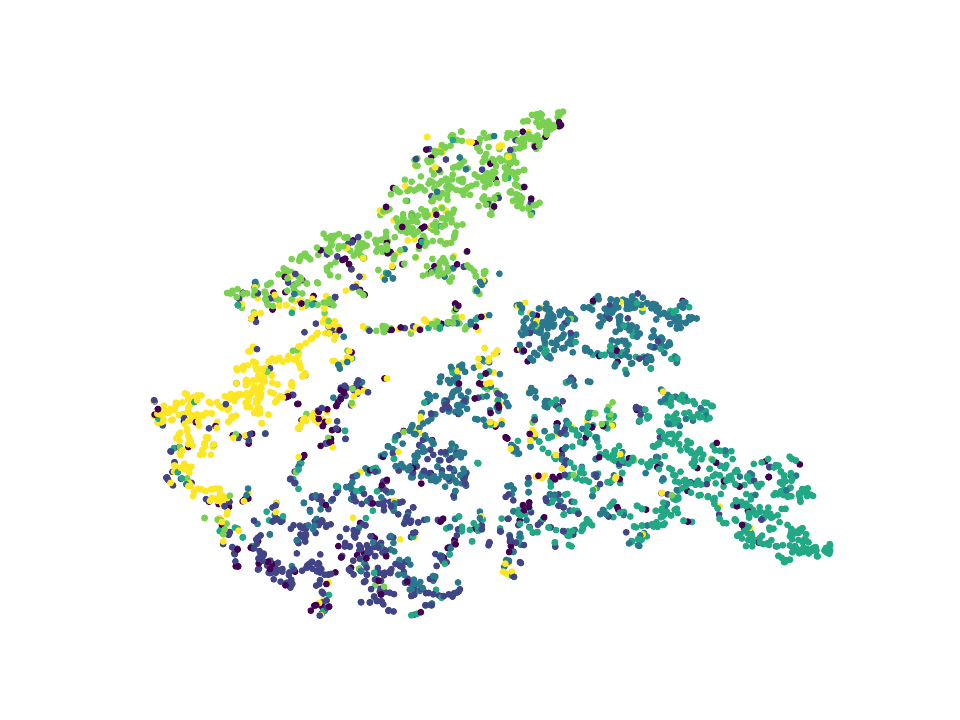}}
    \centerline{\includegraphics[trim=30pt 30pt 30pt 30pt, clip, width=\linewidth]{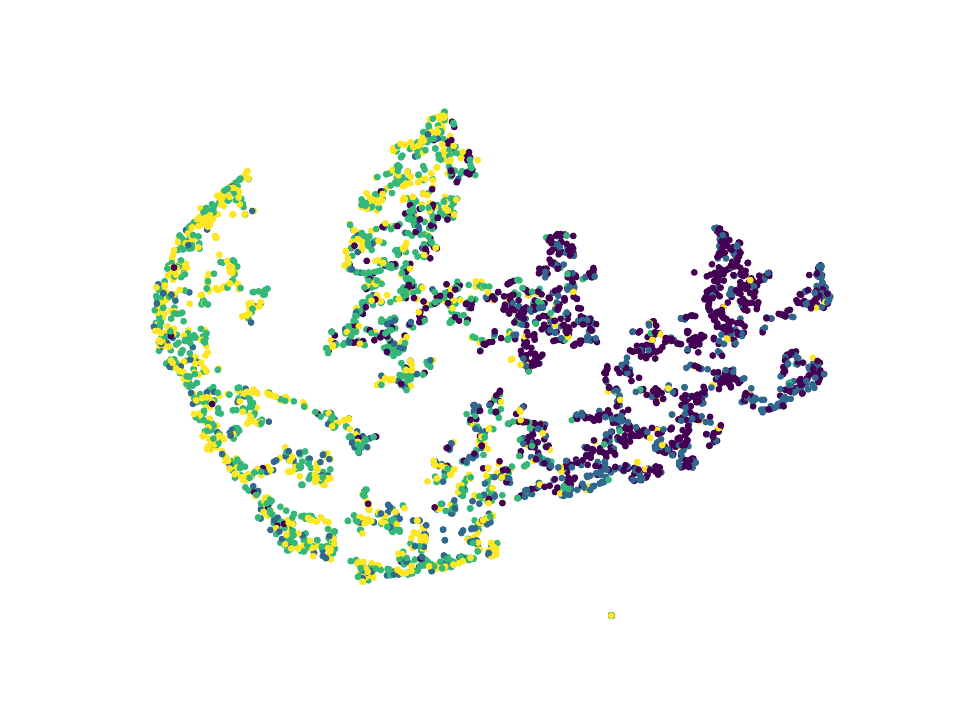}}
  \end{minipage}
    \label{fig:visual_sdcn}
  }
  \hfill
  \subfigure[DFCN]{
  \begin{minipage}{0.1\textwidth}
    \centerline{\includegraphics[trim=30pt 30pt 30pt 30pt, clip, width=\linewidth]{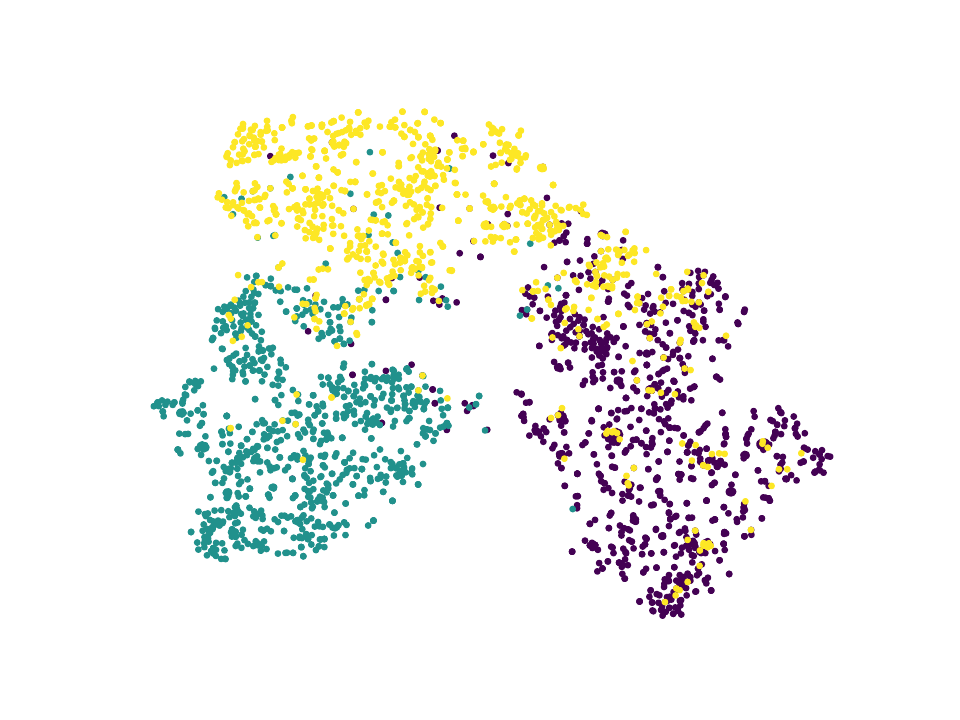}}
    \centerline{\includegraphics[trim=30pt 30pt 30pt 30pt, clip, width=\linewidth]{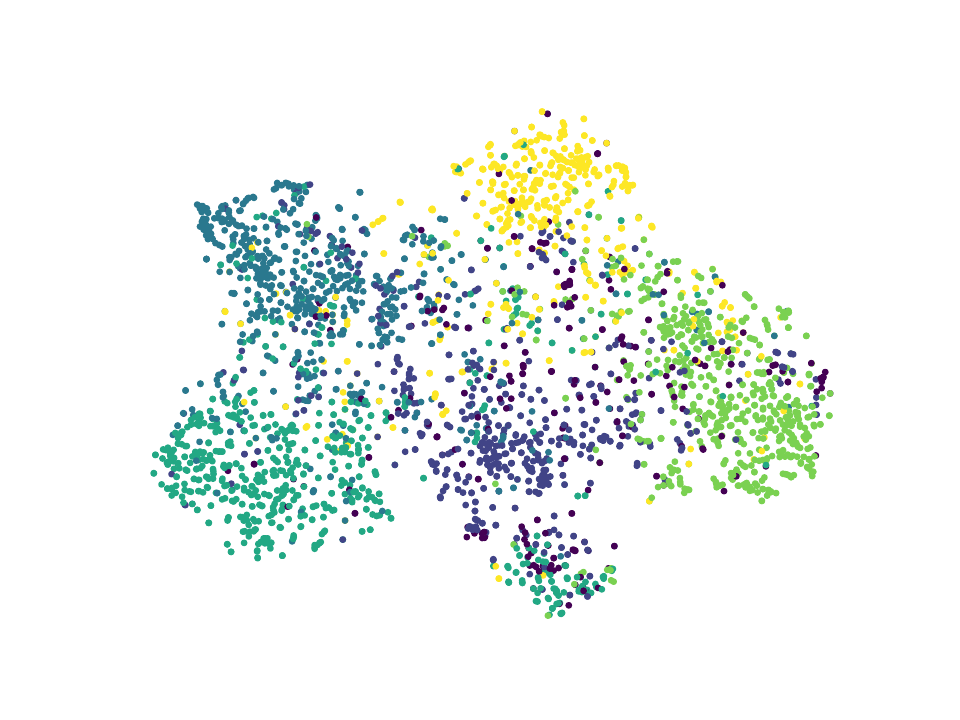}}
    \centerline{\includegraphics[trim=30pt 30pt 30pt 30pt, clip, width=\linewidth]{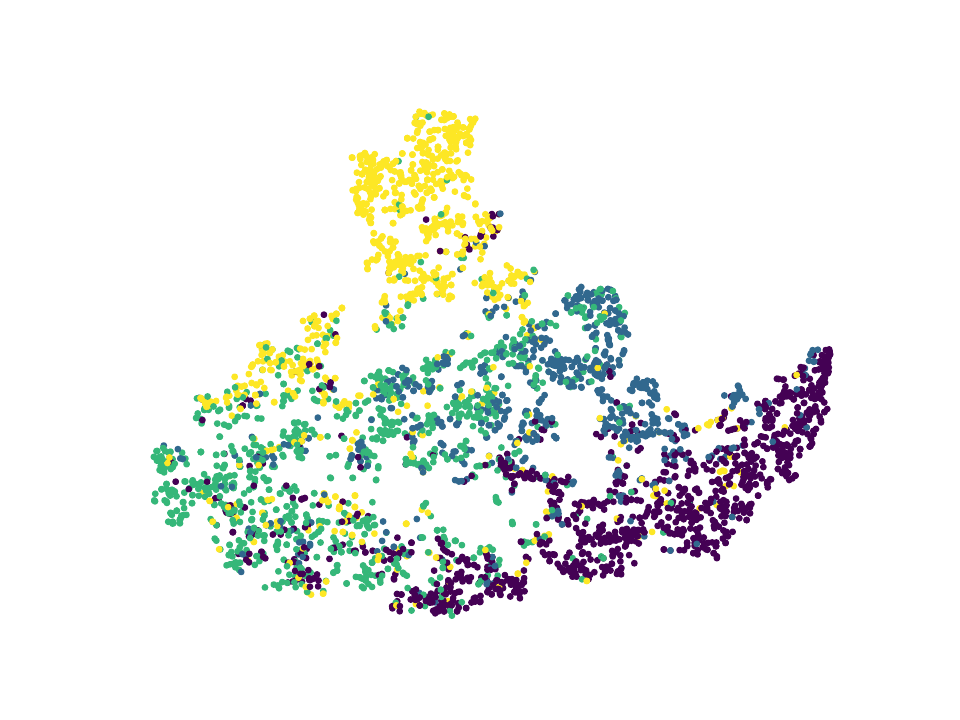}}
  \end{minipage}    
    \label{fig:visual_dfcn}
  }
  \hfill
  \subfigure[DCRN]{
  \begin{minipage}{0.1\textwidth}
    \centerline{\includegraphics[trim=30pt 30pt 30pt 30pt, clip, width=\linewidth]{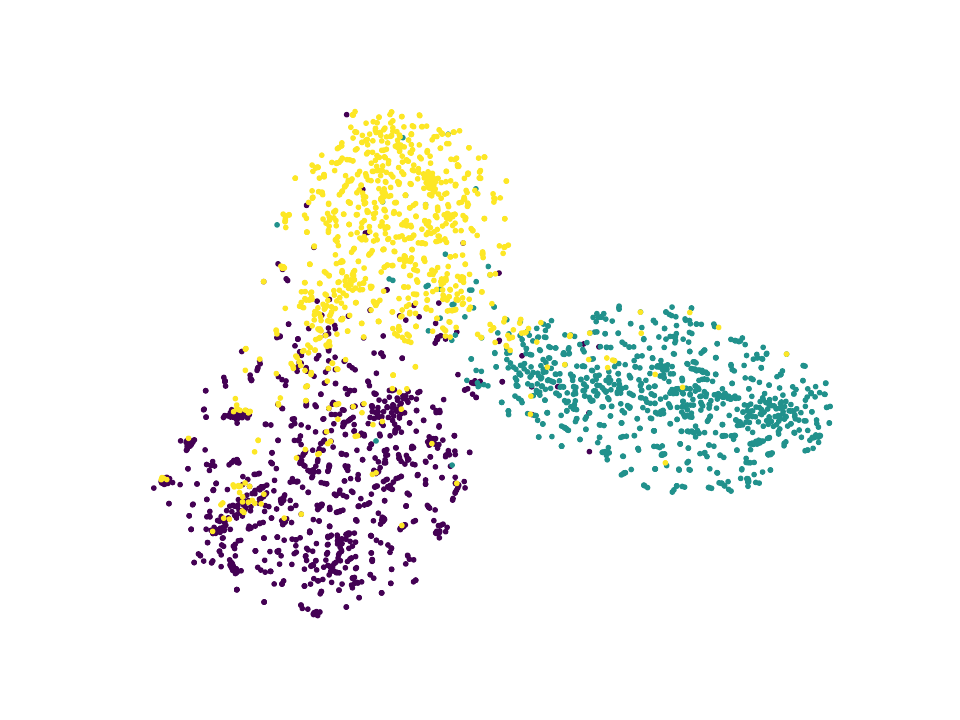}}
    \centerline{\includegraphics[trim=30pt 30pt 30pt 30pt, clip, width=\linewidth]{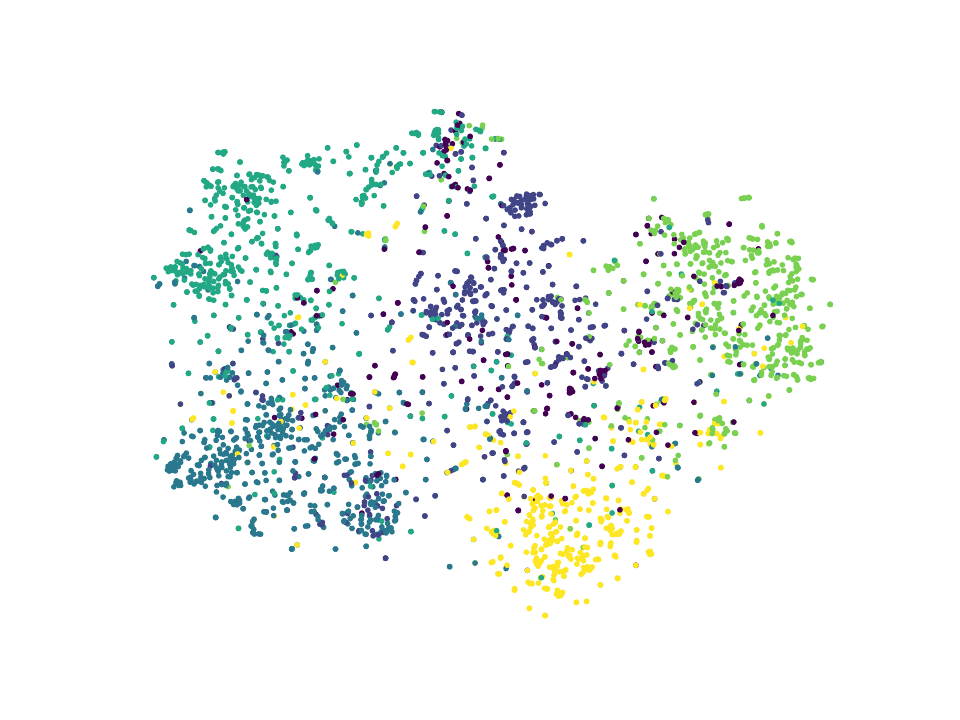}}
    \centerline{\includegraphics[trim=30pt 30pt 30pt 30pt, clip, width=\linewidth]{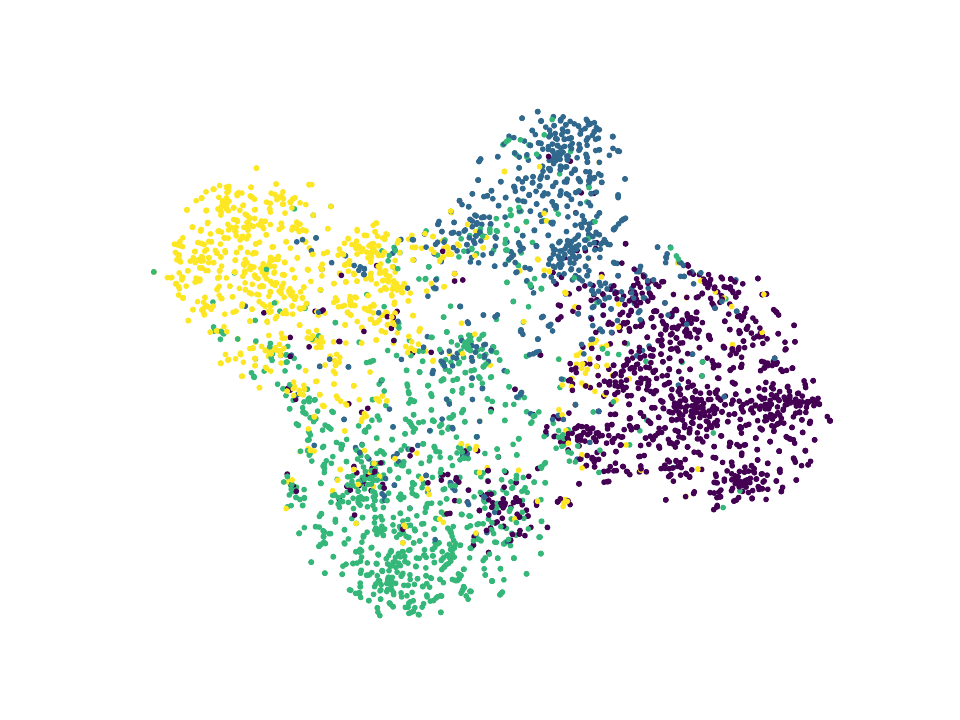}}
  \end{minipage}
    \label{fig:visual_dcrn}
  }
\hfill
  \subfigure[AGC-DRR]{
  \begin{minipage}{0.1\textwidth}
    \centerline{\includegraphics[trim=30pt 30pt 30pt 30pt, clip, width=\linewidth]{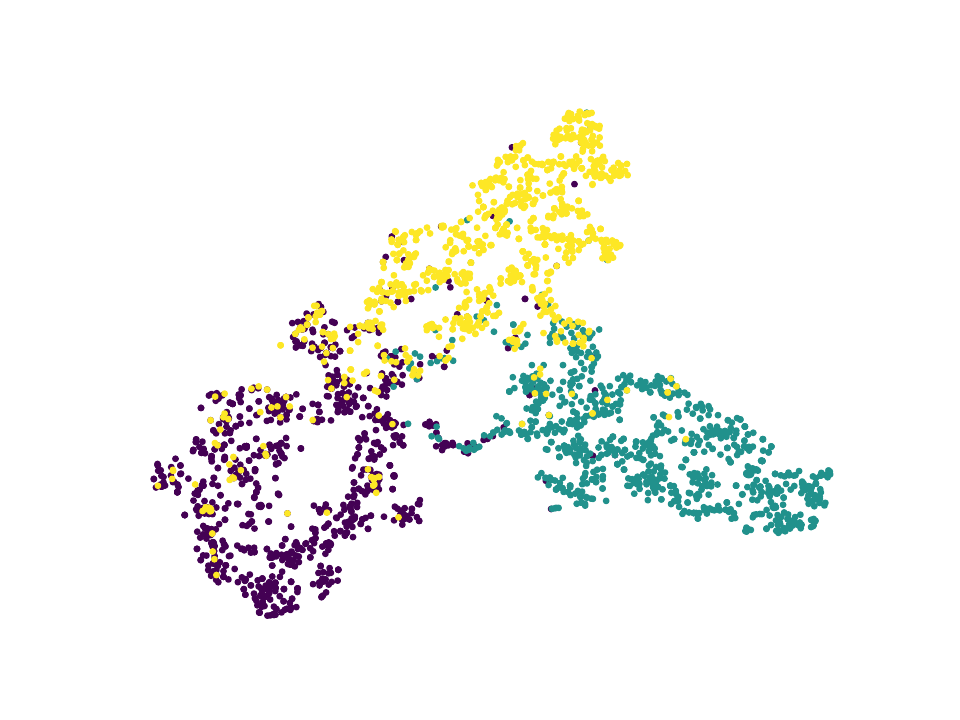}}
    \centerline{\includegraphics[trim=30pt 30pt 30pt 30pt, clip, width=\linewidth]{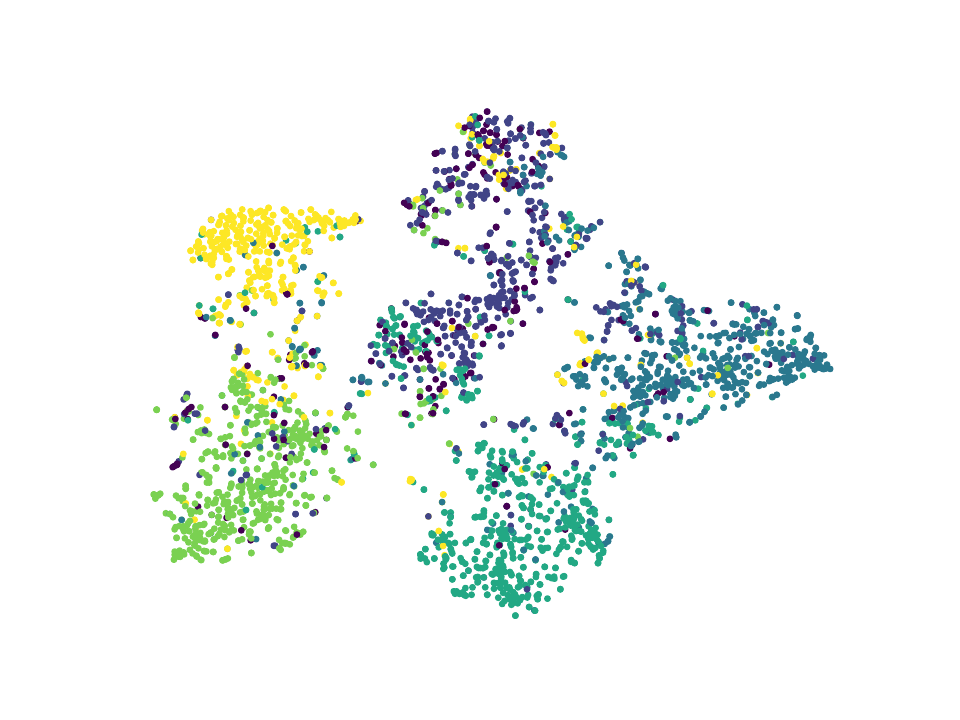}}
    \centerline{\includegraphics[trim=30pt 30pt 30pt 30pt, clip, width=\linewidth]{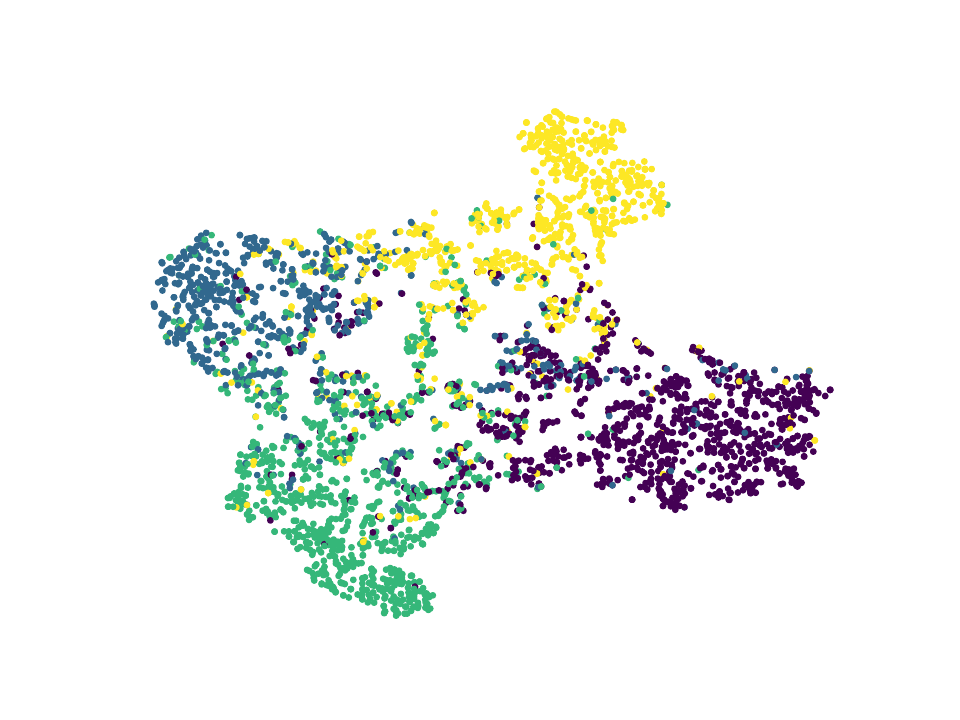}}
  \end{minipage}
    \label{fig:visual_agcdrr}
  }
  \hfill
  \subfigure[CCGC]{
  \begin{minipage}{0.1\textwidth}
    \centerline{\includegraphics[trim=30pt 30pt 30pt 30pt, clip, width=\linewidth]{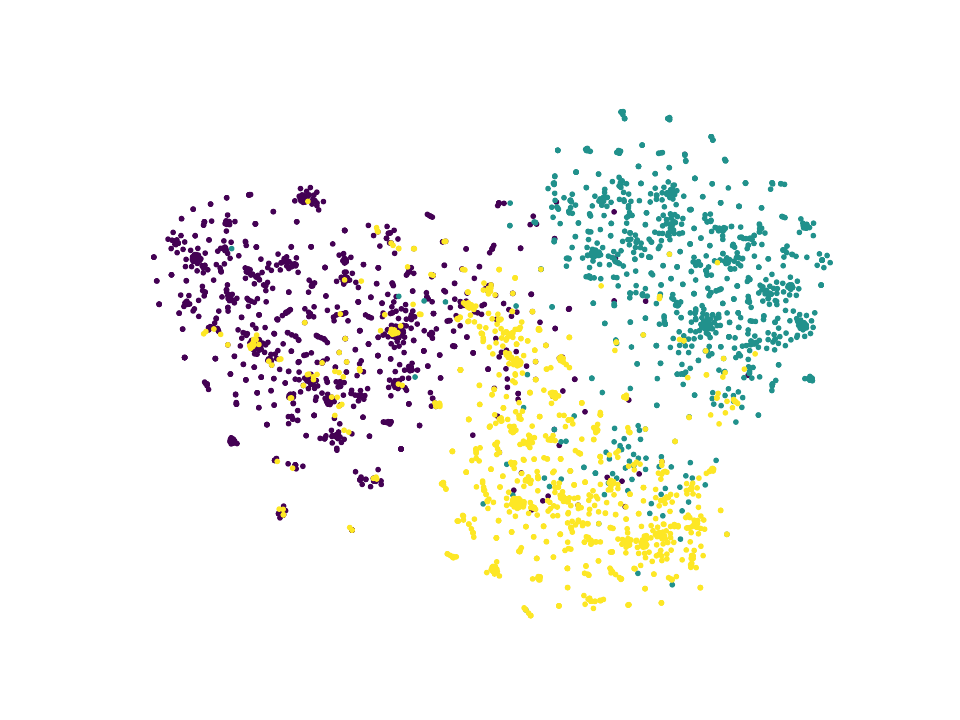}}
    \centerline{\includegraphics[trim=30pt 30pt 30pt 30pt, clip, width=\linewidth]{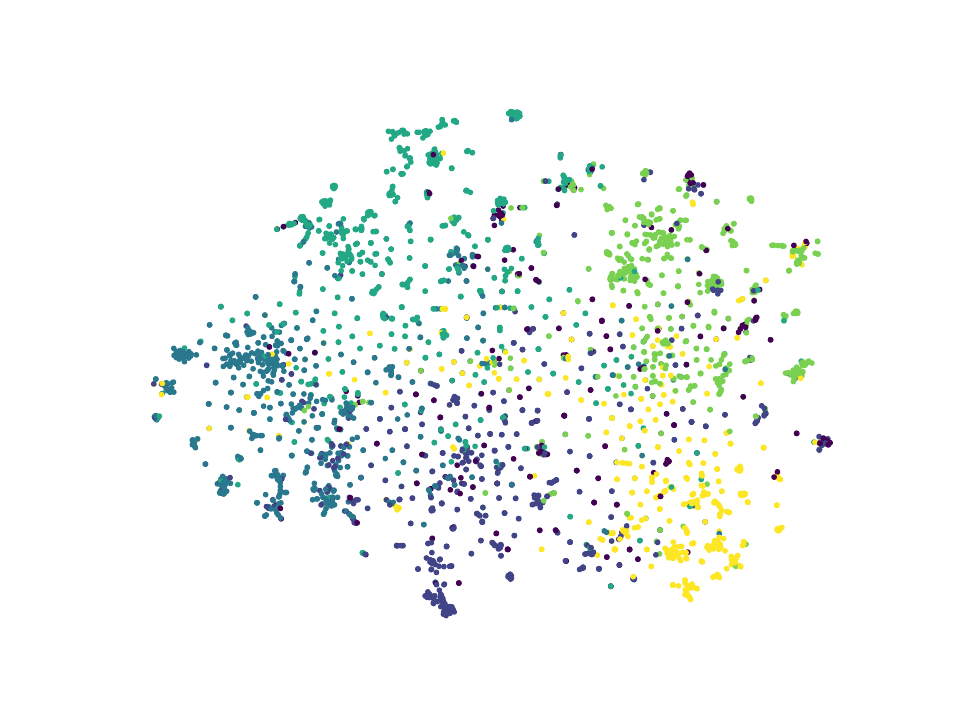}}
    \centerline{\includegraphics[trim=30pt 30pt 30pt 30pt, clip, width=\linewidth]{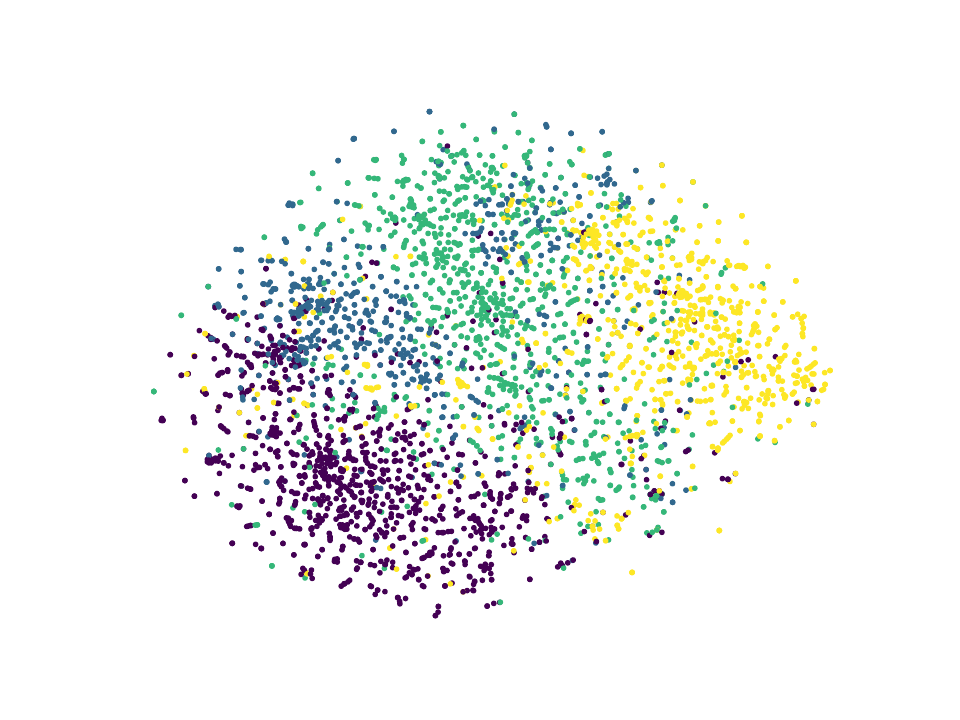}}
  \end{minipage}
    \label{fig:visual_ccgc}
  }
  \hfill
  \subfigure[HSAN]{
  \begin{minipage}{0.1\textwidth}
    \centerline{\includegraphics[trim=30pt 30pt 30pt 30pt, clip, width=\linewidth]{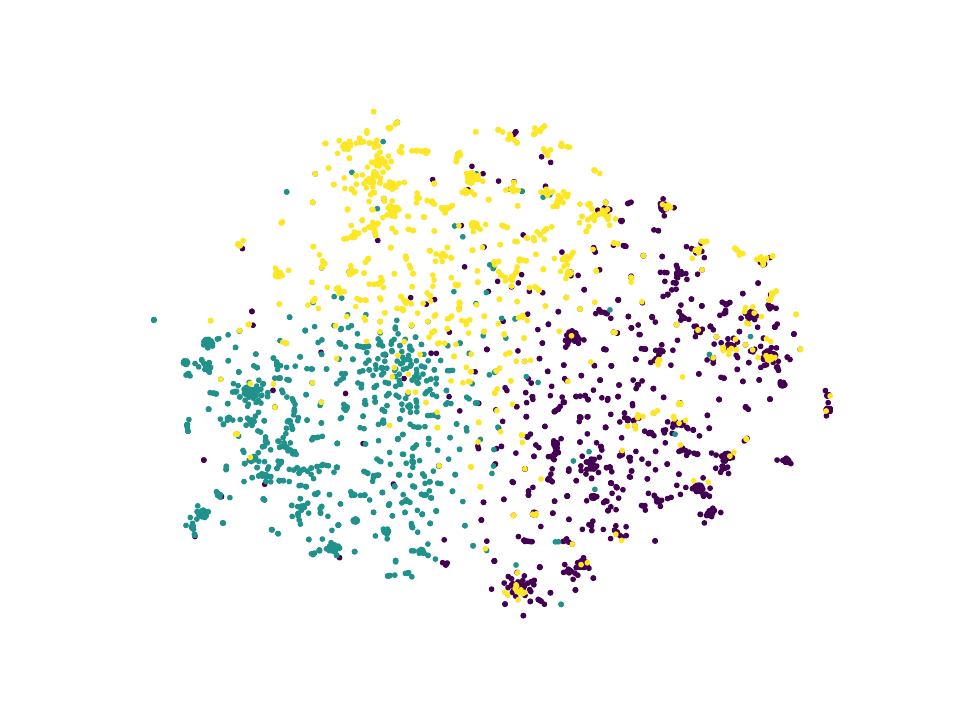}}
    \centerline{\includegraphics[trim=30pt 30pt 30pt 30pt, clip, width=\linewidth]{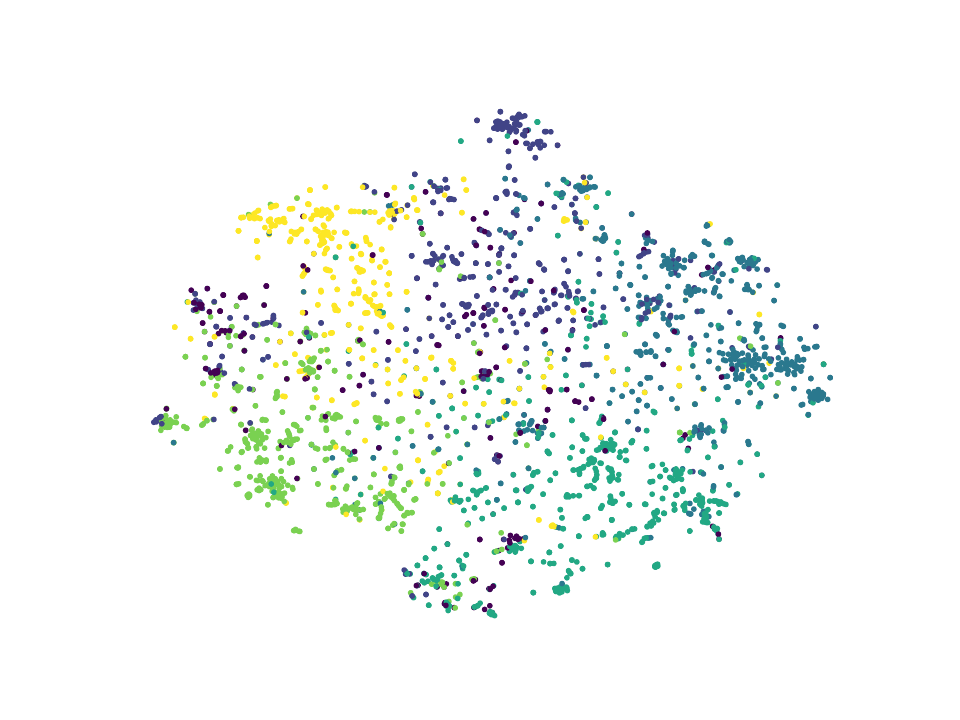}}
    \centerline{\includegraphics[trim=30pt 30pt 30pt 30pt, clip, width=\linewidth]{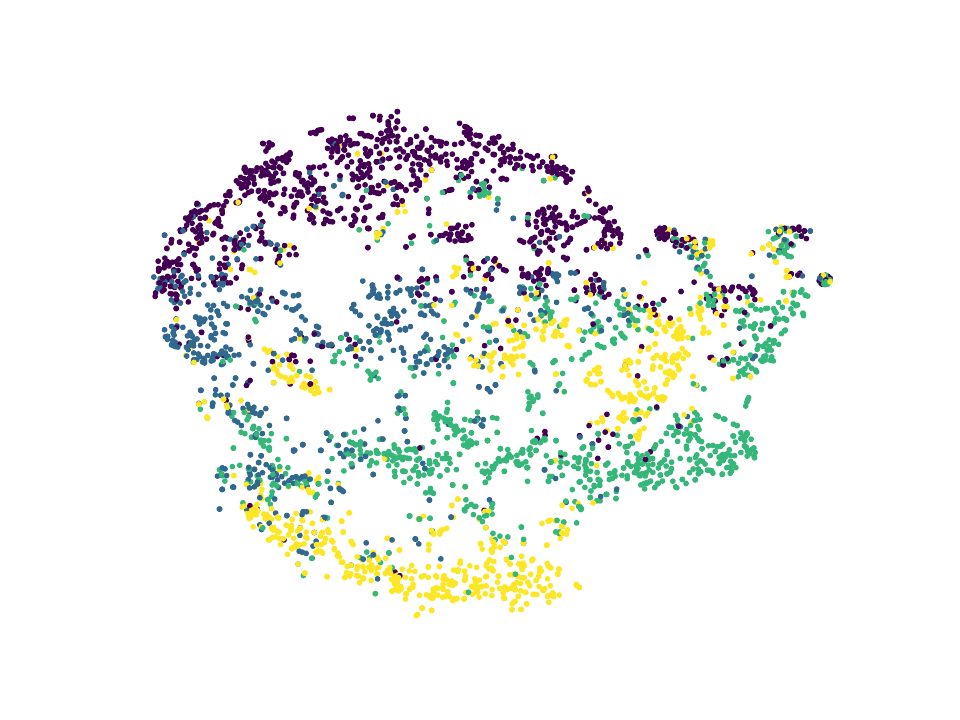}}
  \end{minipage}
    \label{fig:visual_hsan}
  }
  \hfill
  \subfigure[Ours]{
  \begin{minipage}{0.1\textwidth}
    \centerline{\includegraphics[trim=30pt 30pt 30pt 30pt, clip, width=\linewidth]{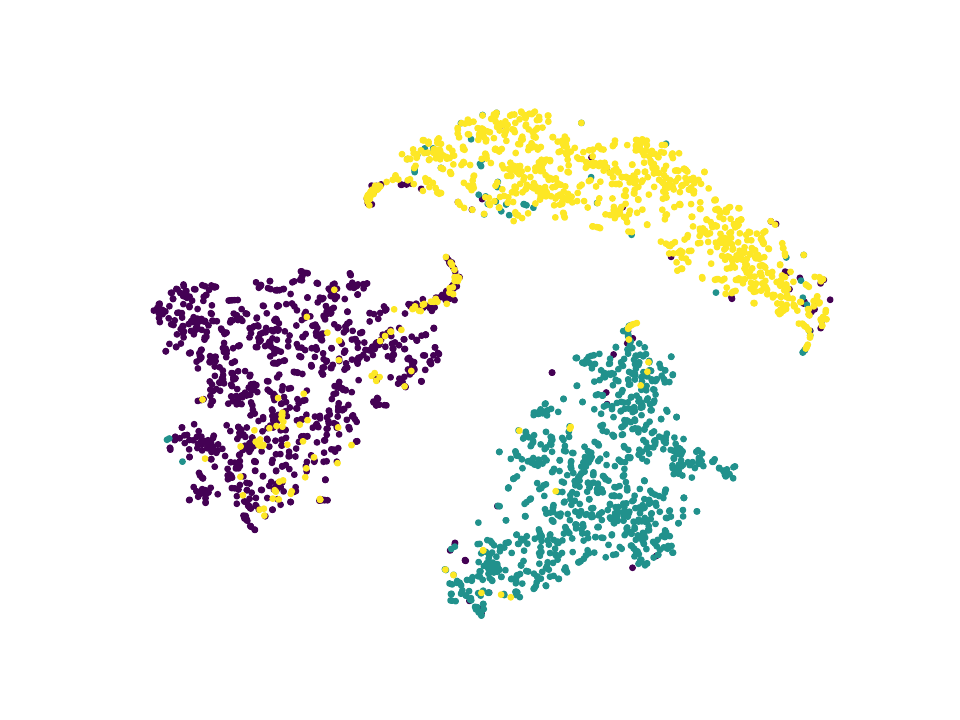}}
    \centerline{\includegraphics[trim=30pt 30pt 30pt 30pt, clip, width=\linewidth]{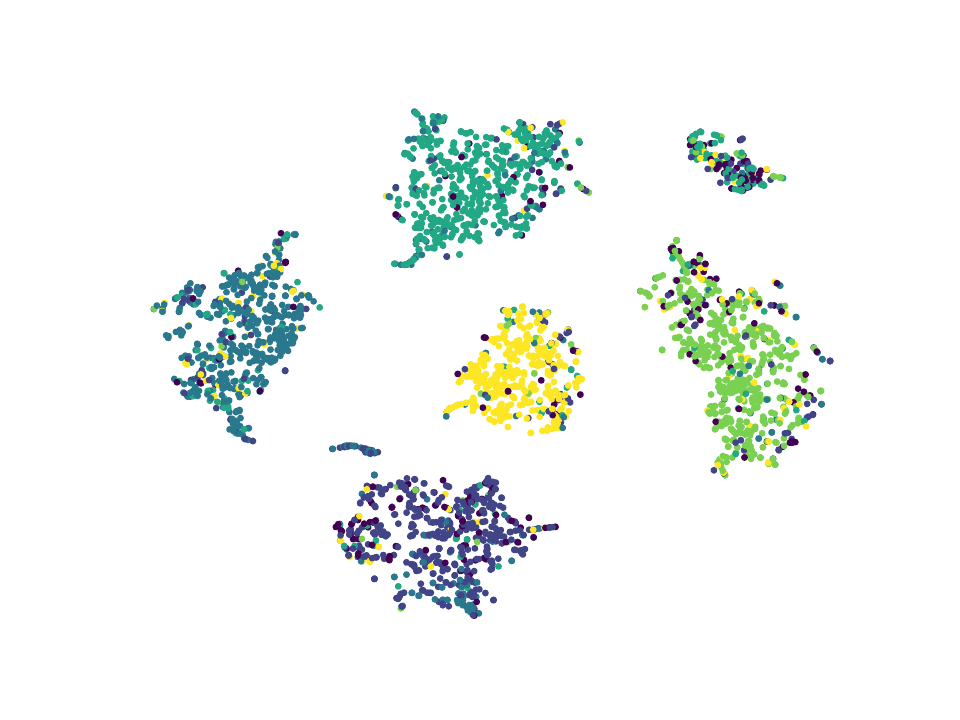}}
    \centerline{\includegraphics[trim=30pt 30pt 30pt 30pt, clip, width=\linewidth]{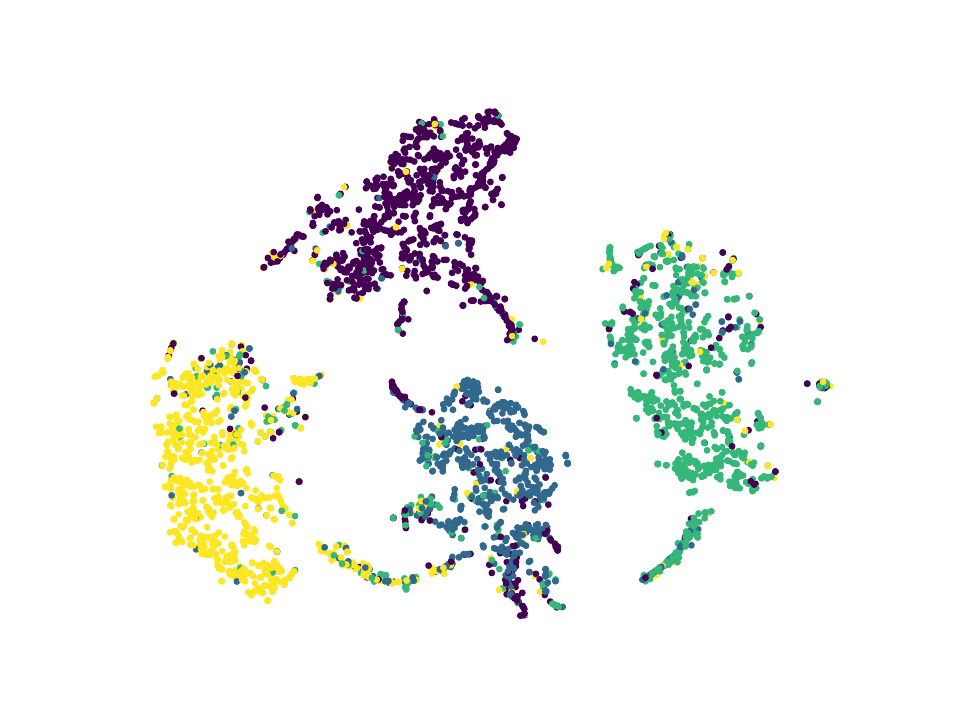}}
  \end{minipage}
    \label{fig:visual_ours}
  }
  \caption{2D $t$-SNE visualization of eight methods. From the first row to the third row, the datasets are ACM, CITE, DBLP.}
  \label{fig:visualization}
\end{figure*}
\begin{figure}[t]
  \centering
  \subfigure[Ideal ACM]{
    \includegraphics[width=0.2\columnwidth]{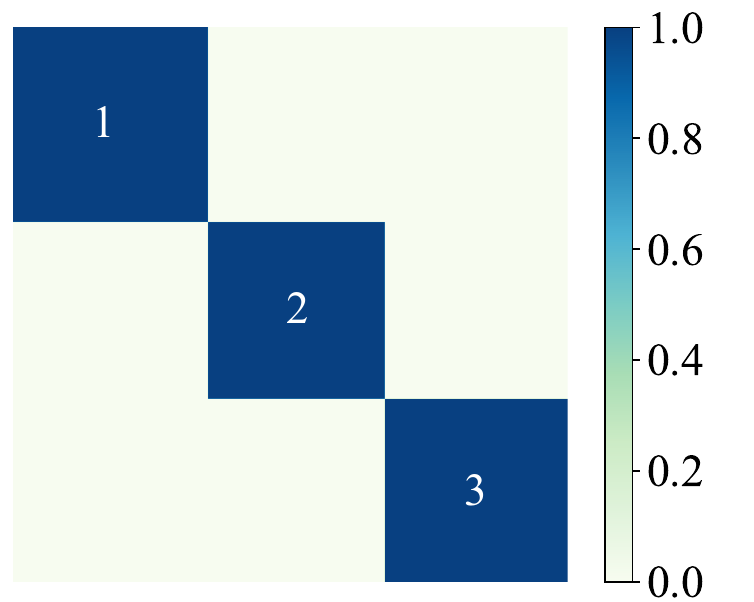}
    \label{fig:acm_g}
  }
  \hfill
    \subfigure[\SynC{} ACM]{
    \includegraphics[width=0.2\columnwidth]{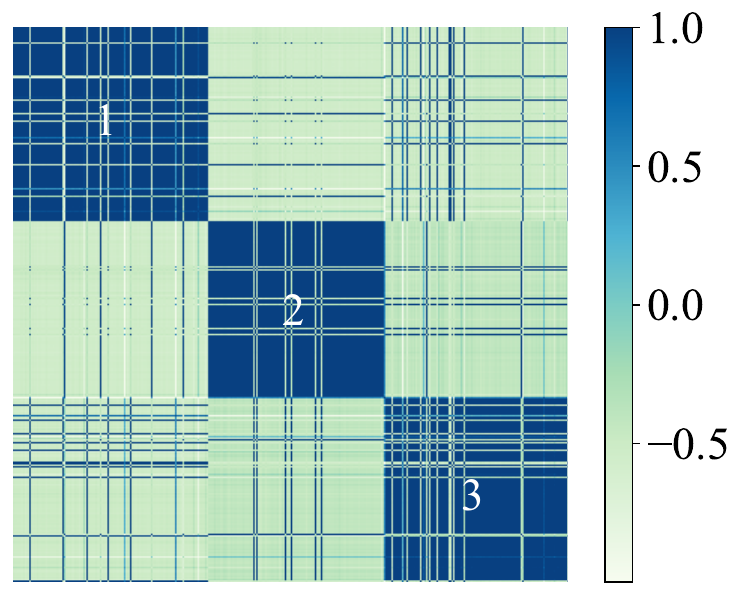}
    \label{fig:acm_e}
  }
    \hfill
  \subfigure[Ideal DBLP]{
    \includegraphics[width=0.2\columnwidth]{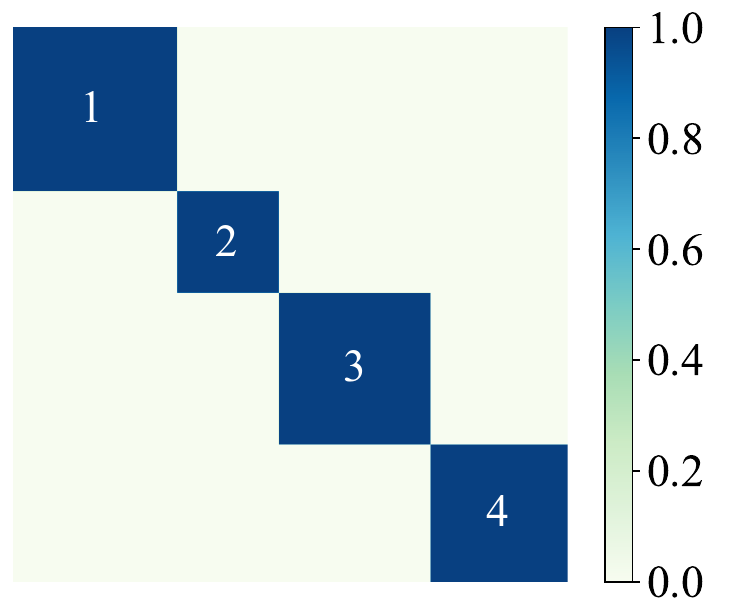}
    \label{fig:dblp_g}
  }
  \hfill
    \subfigure[\SynC{} DBLP]{
    \includegraphics[width=0.2\columnwidth]{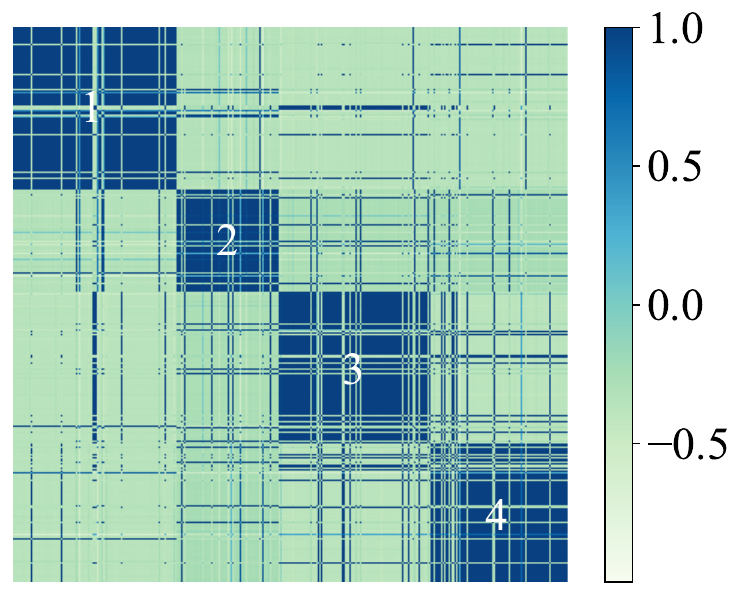}
    \label{fig:dblp_e}
  }

  \subfigure[Ideal CITE]{
    \includegraphics[width=0.2\columnwidth]{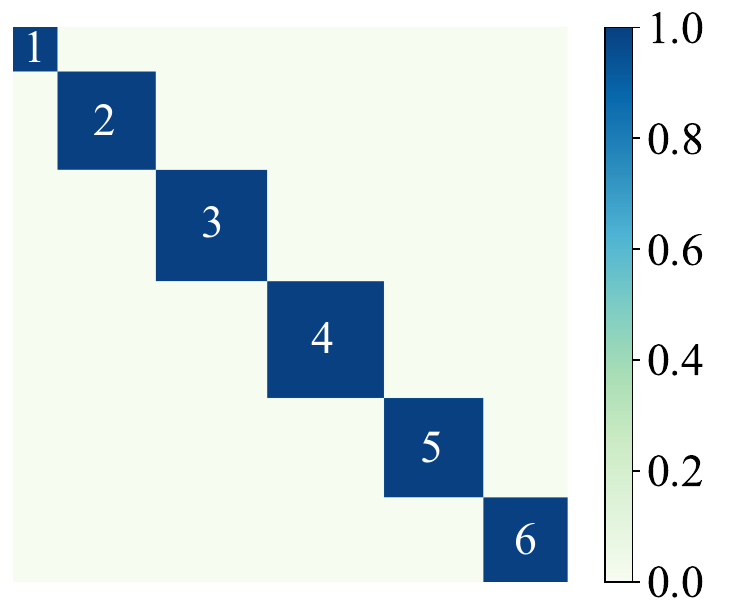}
    \label{fig:cite_g}
  }
  \hfill
    \subfigure[\SynC{} CITE]{
    \includegraphics[width=0.2\columnwidth]{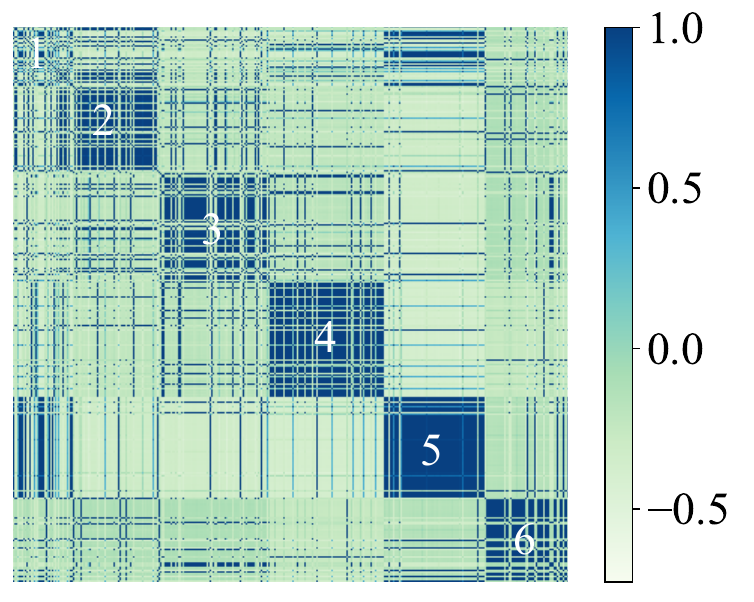}
    \label{fig:cite_e}
  }
  \hfill
  \subfigure[Ideal AMAP]{
    \includegraphics[width=0.2\columnwidth]{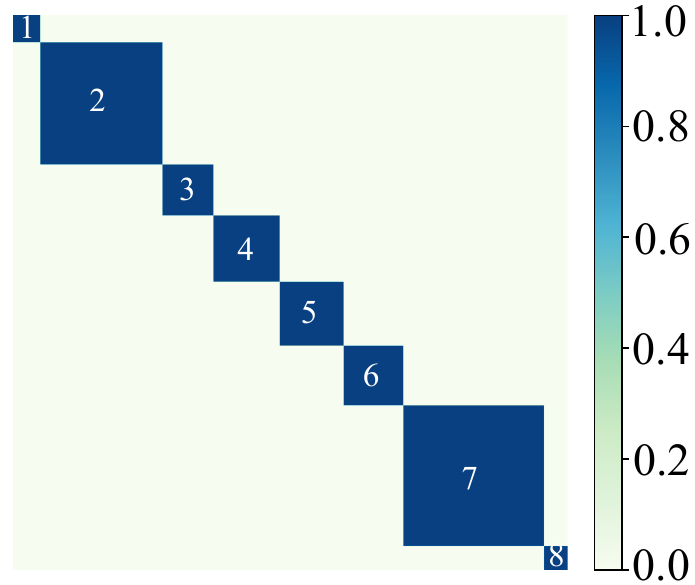}
    \label{fig:amap_g}
  }
  \hfill
    \subfigure[\SynC{} AMAP]{
    \includegraphics[width=0.2\columnwidth]{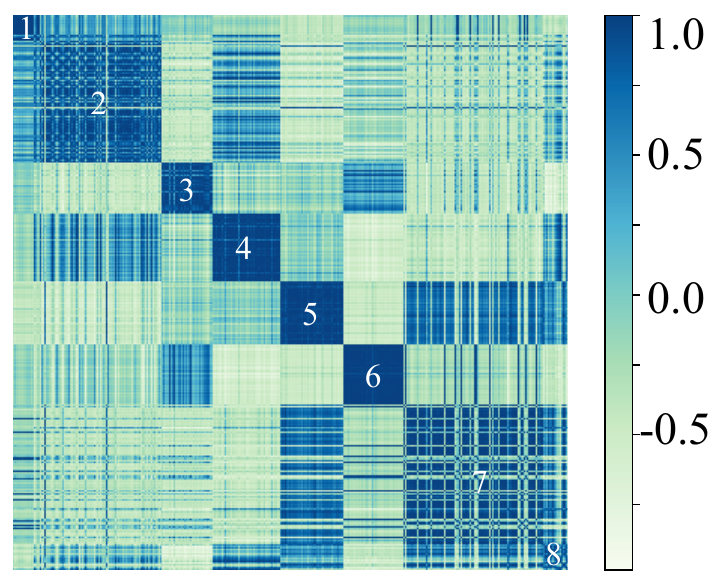}
    \label{fig:amap_e}
  }
  \caption{Ideal embeddings and \SynC{} embeddings similarity heat maps.}
  \label{fig:embeddings}
\end{figure}
\begin{figure}[t]
  \centering
  \subfigure[]{
    \includegraphics[width=0.29\columnwidth]{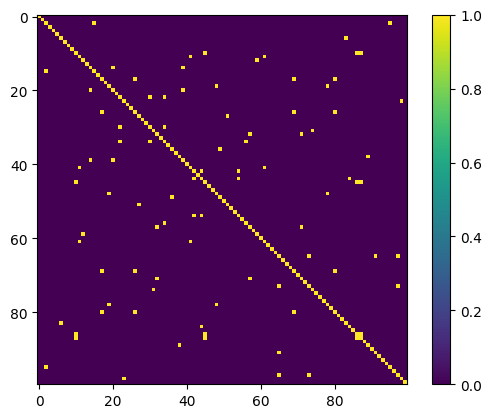}
    \label{fig:add_origin}
  }
  \hfill
  \subfigure[]{
    \includegraphics[width=0.29\columnwidth]{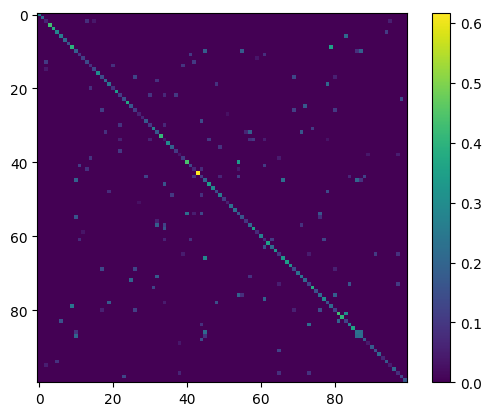}
    \label{fig:add_augmented}
  }
  \hfill
  \subfigure[]{
    \includegraphics[width=0.29\columnwidth]{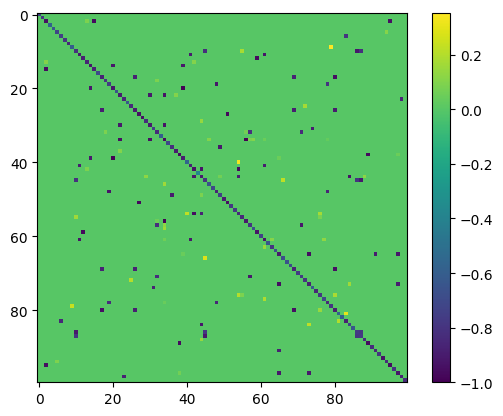}
    \label{fig:add_diff}
  }
  
  \subfigure[]{
    \includegraphics[width=0.29\columnwidth]{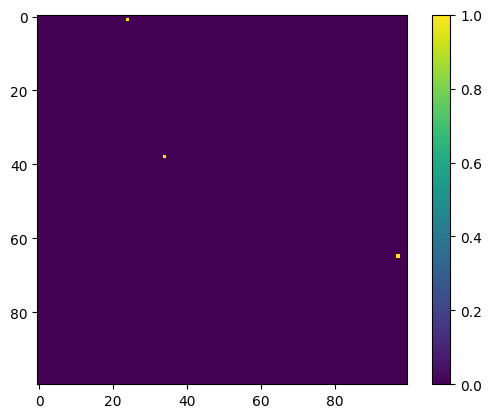}
    \label{fig:remove_origin}
  }
  \hfill
  \subfigure[]{
    \includegraphics[width=0.29\columnwidth]{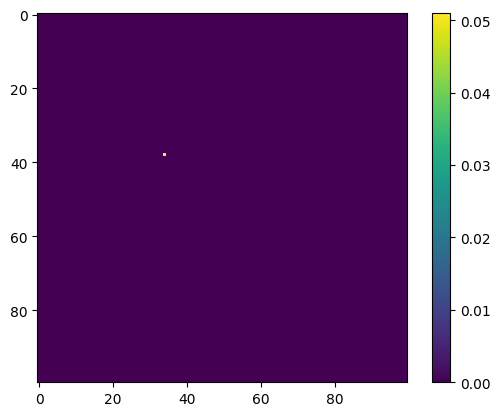}
    \label{fig:remove_augmented}
  }
  \hfill
  \subfigure[]{
    \includegraphics[width=0.29\columnwidth]{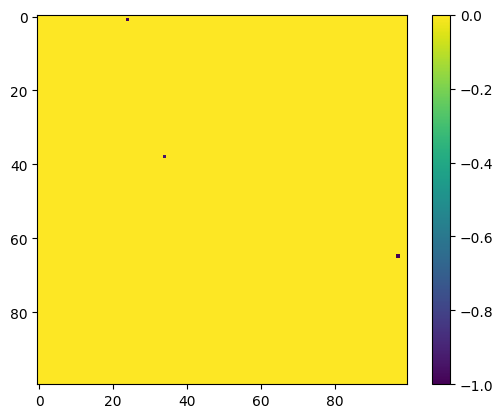}
    \label{fig:remove_diff}
  }
  \caption{\RevisionOne{Original graph and refined graph of nodes from classes 0 (node index: 700-799) and 1 (node index: 1061-1160) on the ACM. (a)-(c) are the adjacency matrices of the graphs between nodes 1061-1160, and (d)-(f) are the adjacency matrices of the graph between nodes 700-799 and nodes 1061-1160. (a) and (d) are the original graphs, and (b) and (d) are refined graphs. (c) is the value of (b) minus (a), and (f) is the value of (e) minus (d), which respectively represent the difference between the two.}}

  \label{fig:graph}
  \end{figure}
\subsection{Visualization Analysis}
$t$-SNE~\cite{van2008visualizing} visualization on the final embeddings was conducted to provide visual evidence of the superior performance of \SynC{}. As depicted in Fig.~\ref{fig:visualization}, \SynC{} outperforms the current state-of-the-art methods, exhibiting superior cluster distribution regarding cohesion and discriminability attributed to the synergistic interaction of representation learning and structure augmentation. 

Similarity heat maps of embeddings were also drawn to visualize the high-quality embeddings learned by our method, which have both strong cohesion and discriminability. As shown in Fig.~\ref{fig:embeddings}, \SynC{} effectively identifies significant clusters and learns cluster-consistent features. At the same time, the features acquired from distinct clusters using our approach exhibit a nearly total dissimilarity. Nevertheless, there are still some challenging nodes that are incorrectly categorized. 

\RevisionOne{Fig.~\ref{fig:graph} demonstrates a comparative visualization between the original graph and the refined graph processed by our proposed method. For this analysis, we selected 100 nodes from class 0 (indices: 700-799) and 100 nodes from class 1 (indices: 1061-1160) in the ACM dataset, specifically examining both intra-class connections within class 1 and inter-class edges between classes 0 and 1. As evidenced in Figs.~\ref{fig:add_origin}-\ref{fig:add_diff}, our method effectively enhances intra-class connectivity while employing edge weighting to quantitatively differentiate connection importance, thereby improving embedding cohesiveness. Furthermore, Figs.~\ref{fig:remove_origin}-\ref{fig:remove_diff} confirm our method's capability in significantly reducing heterophilic connections, consequently enhancing the discriminability of the learned embeddings.}
\subsection{Convergence Analysis}
\begin{figure}[t]
  \centering
  \subfigure[ACM]{
    \includegraphics[trim=0pt 0pt 0pt 30pt, clip, width=0.45\columnwidth]{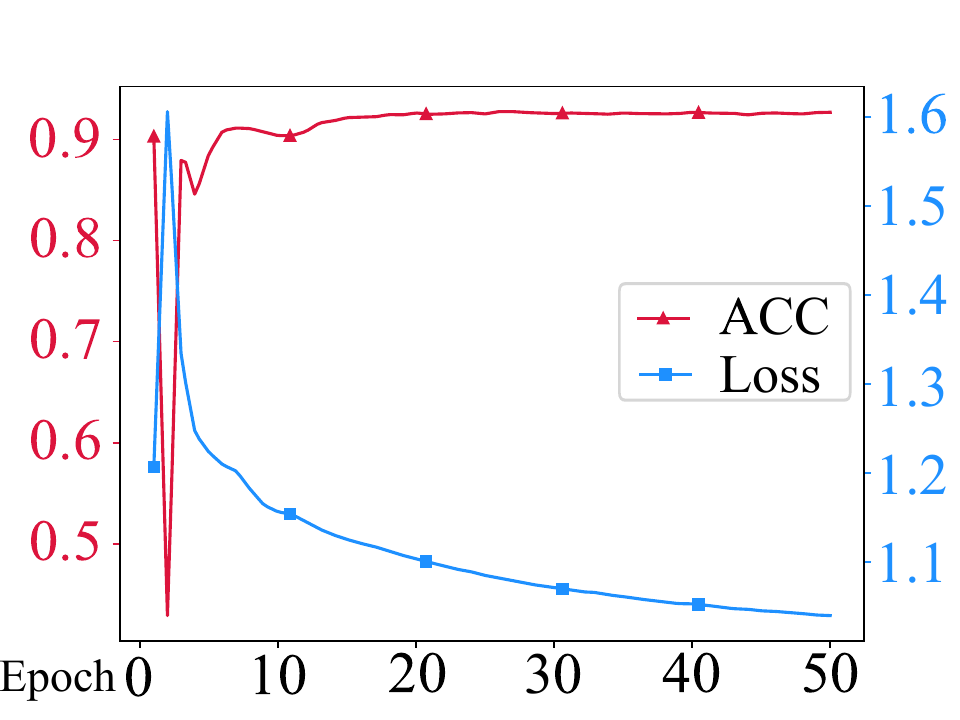}
    \label{fig:acc_loss_acm}
  }
  \hfill
  \subfigure[CITE]{
    \includegraphics[trim=0pt 0pt 0pt 30pt, clip, width=0.45\columnwidth]{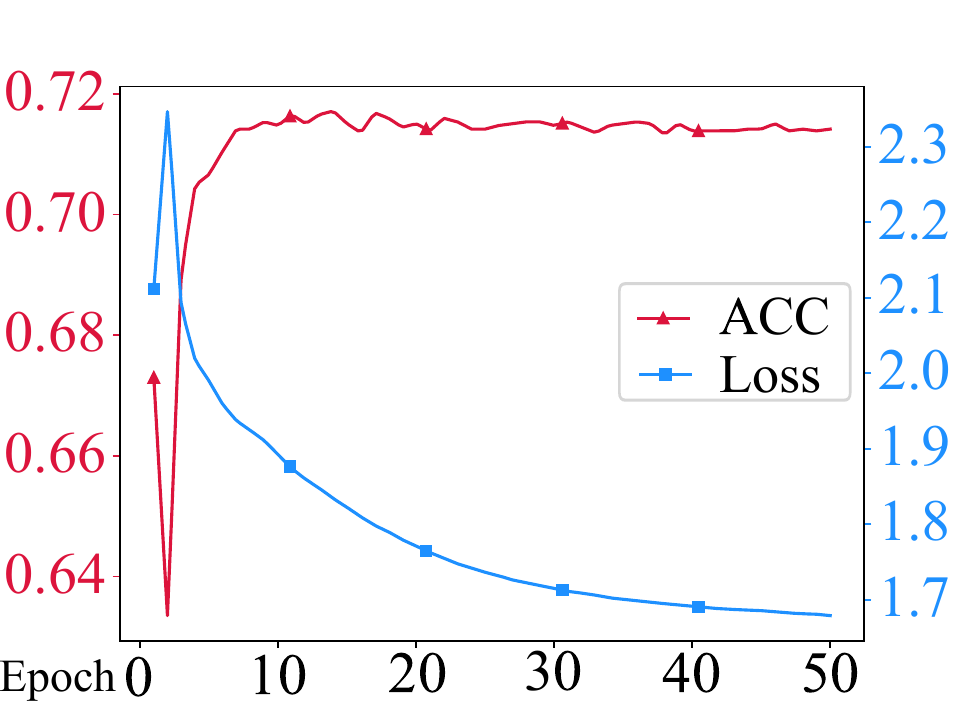}
    \label{fig:acc_loss_cite}
  }
  \caption{Convergence effect on different datasets.}
  \label{fig:acc_loss}
\end{figure}
To demonstrate the convergence of \SynC{}, we conducted experiments on two datasets. We tracked the variations in loss and accuracy throughout the iterations, which are depicted in Fig. \ref{fig:acc_loss}. \SynC{} consistently converged on all datasets, reaching optimal performance within around 20 epochs. During the second iteration, we observed a sharp increase in the loss value and a decline in the accuracy. \RevisionOne{It can be primarily caused by the difference between the loss used in the pre-training stage and that in the training stage. Due to the initialization with pre-trained parameters, the model has a lower loss in the initial stage of training. However, during the training process, as the graph is modified and features are re-aggregated, the discriminative ability of the embeddings is compromised, leading to a sharp increase in clustering loss. Nevertheless, as the synergistic interaction of representation learning and structure augmentation works gradually, this phenomenon disappears, and the model ultimately demonstrates excellent convergence.}

\begin{table}[t]
\caption{\RevisionOne{Comparison of time and space consumption on the ACM.}}
\centering
\resizebox{\columnwidth}{!}{
\begin{tabular}{c|ccccc}
\Xhline{1pt}
\diagbox{\textbf{Methods}}{\textbf{Metrics}}
&\textbf{SDCN} & \textbf{DCRN} & \textbf{AGC-DRR} & \textbf{HSAN} & \RevisionOne{\textbf{\SynC{}}} \\
\Xhline{0.5pt}

\makecell{Pre-training Time (s)} &6.38&38.94&\textbf{0}&\textbf{0}&2.7\textsuperscript{*} \\
\makecell{Training Time (s)} &14.71\textsuperscript{*}&147.10&2165.31&OOM&\RevisionOne{\textbf{8.25}} \\
\makecell{Iterations} &\textbf{50}&400\textsuperscript{*}&400\textsuperscript{*}&400\textsuperscript{*}&\textbf{50} \\
\makecell{Training  Speed (it/s)} &3.40\textsuperscript{*}&2.72&0.18&OOM&\RevisionOne{\textbf{6.06}} \\
\makecell{Params (M)}&6.62&\textbf{0.60}&1.60&10.58&1.09\textsuperscript{*} \\
\makecell{Max GPU  Memory (MB)}&507.71\textsuperscript{*}&1189.40&\textbf{348.08}&1935.58&\RevisionOne{602.06} \\
\makecell{Average Rank} &2\textsuperscript{*}&4&2\textsuperscript{*}&5&\textbf{1} \\

\Xhline{1pt}
\end{tabular}
}
\label{table:time_comp}
\end{table}

\subsection{Time and Space Consumption Analysis}
Performance experiments were conducted to compare the comprehensive performance of \SynC{} with other state-of-the-art methods on the ACM dataset. The evaluation considered various performance indicators, including pre-training time, training time, iterations, training speed (iterations per second), parameter count (Params, in millions), maximum GPU memory (in MB), and average rank. ``OOM" denotes ``out of memory" when training on a GTX 1050. The results in Table \ref{table:time_comp} show that \SynC{} achieves the best overall performance.
\RevisionOne{\subsection{Robustness Analysis}}
\label{sec:robustness}
\begin{table}[t]
\caption{\RevisionOne{Clustering performance of \SynC{} with random perturbations on the dataset ACM and DBLP.}}
\centering
\label{tab:perturbation}
\resizebox{\columnwidth}{!}{
\begin{tabular}{cccccc}\toprule
\textbf{Noisy Type}                    & \textbf{Dataset}               & \textbf{Propotion} & \textbf{0} & \textbf{0.1} & \textbf{0.2}  \\ \midrule
\multirow{4}{*}{\textbf{Mask Feature}} & \multirow{2}{*}{\textbf{ACM}}  & \textbf{TIGAE}    & 90.41      & 90.58        & 89.85             \\
                                       &                                & \textbf{\SynC{}}       & 92.74±0.04 & 92.52±0.19   & 91.35±0.54     \\
                                       & \multirow{2}{*}{\textbf{DBLP}} & \textbf{TIGAE}    & 78.21      & 76.61        & 74.74          \\
                                       &                                & \textbf{\SynC{}}       & 83.48±0.15 & 75.59±2.41   & 73.60±3.44     \\ \midrule
\multirow{4}{*}{\textbf{Add Edge}}     & \multirow{2}{*}{\textbf{ACM}}  & \textbf{TIGAE}    & 90.41      & 85.62        & 83.07          \\
                                       &                                & \textbf{\SynC{}}       & 92.74±0.04 & 91.15±0.29   & 90.04±0.22     \\
                                       & \multirow{2}{*}{\textbf{DBLP}} & \textbf{TIGAE}    & 78.21      & 76.83        & 74.12          \\
                                       &                                & \textbf{\SynC{}}       & 83.48±0.15 & 82.37±0.25   & 80.07±1.63    \\ \bottomrule
\end{tabular}}
\end{table}
\RevisionOne{To validate \SynC{}'s capability in handling noise, we conducted experiments on ACM and DBLP datasets by simulating perturbations through random feature masking and random edge addition. The perturbation ratios were set to 0, 0.1, and 0.2, respectively (where 0 indicates no perturbation). For feature masking, the ratio is calculated based on the feature dimension; for edge addition, the ratio is computed relative to the total number of edges. The experimental results in Table \ref{tab:perturbation} demonstrate that while our method's performance shows some degree of degradation with increasing perturbation intensity, \SynC{} exhibits significantly better robustness compared to using a single TIGAE (pretrain). This outcome further confirms the effectiveness of leveraging the reciprocal relationship between graph representation learning and structural augmentation for learning cluster-oriented embeddings.}
\vspace{-0.5pt}
\section{Conclusion}
\label{sec:conclusion}
In this paper, we propose a clustering framework termed Synergistic Deep Graph Clustering Network (\SynC{}), which fully leverages the feature-smoothing capability of GNNs. Besides, we design a Transform Input Graph Auto-Encoder (TIGAE), which utilizes a linear transformation with cosine similarity preserved to indirectly incorporate explicit structural information, addressing the representation collapse issue of GAE. TIGAE optimizes the computational complexity of GAE under certain conditions, resulting in faster computational speed for \SynC{}. The structure fine-tuning strategy improves the generalization of \SynC{}. Our method stands out for its state-of-the-art clustering results, fast execution speed, and high stability. Extensive experiments provide accurate and compelling evidence for the outstanding performance of \SynC{}. \RevisionOne{However, \SynC{} uses full batch training and is designed for small-scale data. In the future, we will investigate how to achieve the synergistic interaction between representation learning and structure augmentation in large-scale graph data.}
\bibliographystyle{IEEEtran.bst}
\bibliography{references}





\end{document}